\documentclass{IEEEtran}
\usepackage{hyperref}       
\usepackage{url}            
\usepackage{booktabs}       
\usepackage{amsfonts}       
\usepackage{nicefrac}       
\usepackage{microtype}      
\usepackage{float}
\usepackage{lipsum}
\usepackage{algpseudocode}
\usepackage[ruled,vlined]{algorithm2e}
\usepackage{xcolor}
\usepackage{epsfig, graphicx, amsmath, amssymb,cite}
\usepackage{lipsum}
\usepackage{mwe}
\newtheorem{example}{Example}

\DeclareMathOperator*{\argmin}{\arg\!\min}
\IEEEoverridecommandlockouts

\begin{document}
\title{\huge Learning-Aided Physical Layer Attacks Against Multicarrier Communications in IoT}
\author{\IEEEauthorblockN{Alireza Nooraiepour, Waheed U. Bajwa and Narayan B. Mandayam\\}
\thanks{The authors are with WINLAB, Department of Electrical and Computer Engineering, Rutgers University, NJ, USA. Emails: {$\{$alinoora,narayan$\}$}@winlab.rutgers.edu, {waheed.bajwa}@rutgers.edu.

This work was supported in part by the NSF under Grant No. ACI-1541069. A preliminary version of this work was presented in 2018 IEEE Military Communications Conference (MILCOM) \cite{OURMILCOM}.
}
    }
\date{}
\maketitle

\begin{abstract}

Internet-of-Things (IoT) devices that are limited in power and processing capabilities are susceptible to physical layer (PHY) spoofing (signal exploitation) attacks owing to their inability to implement a full-blown protocol stack for security. The overwhelming adoption of multicarrier techniques such as orthogonal frequency division multiplexing (OFDM) for the PHY layer makes IoT devices further vulnerable to PHY spoofing attacks. These attacks which aim at injecting bogus/spurious data into the receiver, involve inferring transmission parameters and finding PHY characteristics of the transmitted signals so as to spoof the received signal. Non-contiguous (NC) OFDM systems have been argued to have low probability of exploitation (LPE) characteristics against classic attacks based on cyclostationary analysis, and the corresponding PHY has been deemed to be secure. However, with the advent of machine learning (ML) algorithms, adversaries can devise data-driven attacks to compromise such systems. It is in this vein that PHY spoofing performance of adversaries equipped with supervised and unsupervised ML tools are investigated in this paper. The supervised ML approach is based on estimation/classification utilizing deep neural networks (DNN) while the unsupervised one employs variational autoencoders (VAEs). In particular, VAEs are shown to be capable of learning representations from NC-OFDM signals related to their PHY characteristics such as frequency pattern and modulation scheme, which are useful for PHY spoofing. In addition, a new metric based on the disentanglement principle is proposed to measure the quality of such learned representations. Simulation results demonstrate that the performance of the spoofing adversaries highly depends on the subcarriers' allocation patterns used at the transmitter. Particularly, it is shown that utilizing a random subcarrier occupancy pattern precludes the adversary from spoofing and secures NC-OFDM systems against ML-based attacks.
\end{abstract}

\section{Introduction}
The broadcast nature of radio signal propagation along with standardized transmission schemes and intermittent communications make wireless communication systems extremely vulnerable to interception and spoofing attacks. Specifically, the broadcast
nature of the wireless medium facilitates the reception of radio
signals by any illegitimate receiver as long as it is within the coverage radius of
the transmitter. Further, standardized transmission
and conventional security schemes open up wireless systems to interception and eavesdropping \cite{Intro1}. Additionally, sporadic transmissions of low-cost wireless devices, especially the significantly growing
number of Internet-of-Things (IoT) devices, provide massive
opportunities to adversaries and malicious actors for spoofing attacks. Therefore,
it is of paramount
importance for wireless communication systems to enhance the security mechanisms meant to combat the adversaries, especially in light of the ongoing adoption of IoT systems in industrial applications.

IoT devices that have limited battery
and computational resources may not be able to execute a full-blown protocol stack for security and authentication (non-access stratum (NAS)) purposes \cite{NAS}.
Physical layer (PHY) security has been put forth as a promising solution in this case, which
aims at exploiting the PHY characteristics of the communication links. We
study PHY spoofing attacks in IoT networks enabled with multicarrier communications where
an adversary spoofs a legitimate transmitter by sending bogus data to
its intended receiver. Specifically, we consider an adversary which is equipped with cutting-edge machine learning (ML) tools, and aims at inferring PHY parameters of OFDM/NC-OFDM signals used by the transmitter. Knowledge of these parameters enables the adversary to generate signals similar to those of the transmitter's and perform spoofing attacks. It is in this context that we investigate the robustness of OFDM/NC-OFDM systems against ML-aided PHY spoofing attacks. Our focus is in particular on understanding the corresponding PHY characteristics that can help secure such IoT systems.
\subsection{Relation to prior work}

Current attempts at using PHY characteristics as authentication keys for the message source follow various approaches. One possibility is to assume a pre-shared secret key hidden in the modulation scheme, which is detected by the receiver \cite{preshared,PHY3}. In other keyless transmitter-based methods (also known as wireless fingerprinting), device-specific non-ideal transmission parameters are extracted from the received signal. These are identified as characteristics of the claimed source and then compared with those from previous authenticated messages \cite{shortpaper}. Channel-based authentication algorithms \cite{freq1}, \cite{freq2} compare the channel response estimated from the current message with that estimated from the previous ones by the claimed source, thus actually authenticating the position of the transmitter rather than its identity. However, the attacker in these works is assumed to only use higher-layer identity forging (e.g., spoofing of a MAC address)
and does not try to attack the system by exploiting the underlying PHY characteristics of the
signals. The latter approach is referred to as PHY spoofing, which is an important attack that has to be studied since it has been shown to be able to compromise OFDM systems, which are widely used in IoT standards \cite{NBIOT1}. Therefore, we investigate the performance of powerful attackers enabled with state-of-the-art ML algorithms in PHY spoofing, and propose exploiting PHY characteristics to combat such attacks.

As mentioned, PHY spoofing is shown to be able to compromise OFDM systems. NC-OFDM systems in which transmissions take place over a subset of subcarriers are shown to be capable of circumventing this impediment \cite{NCOFDM}. NC-OFDM systems are also capable of efficiently utilizing the fragmented spectrum and improving spectral efficiency. The authors in \cite{RatenshMilCom,OURMILCOM} examined the low
probability of exploitation (LPE) characteristics of NC-OFDM
transmissions, assuming that an adversary is using cyclostationary
analysis \cite{CAF1,CAF2,CAF3} to infer transmission parameters. 
In \cite{RatenshMilCom}, the authors showed that
the cyclostationary analysis is extremely challenging to do for
most choices of NC-OFDM transmission parameters. Therefore, NC-OFDM systems may be deemed to be secure against PHY spoofing. However, it is not clear if this is still the
case when an adversary utilizes powerful data-driven tools like ML algorithms for spoofing.

Particularly, a wide range of ML tools have recently received a lot of attention among communication researchers for solving analytically intractable problems. In particular, the authors in \cite{Deepsig} proposed an end-to-end learning of communication systems based on deep neural networks (DNNs). They optimize transmitter and receiver jointly without considering the classical communication and signal processing blocks, including channel encoder and modulator. In \cite{DeepforChannel}, the authors showed that deep learning techniques are very promising in scenarios where the channel is too complex to be described analytically. The authors in \cite{overtheair} demonstrated how a DNN-based system can communicate over-the-air without the need for any conventional signal processing blocks. Moreover, securing a point-to-point communication system against a DNN-based attacker trying to determine the modulation scheme is considered in \cite{FukingGunduz}.

\subsection{Our contributions}
We investigate the resilience of OFDM/NC-OFDM systems against PHY spoofing, assuming the adversary is equipped with ML tools. We assume the adversary continuously listens to the transmitting terminal, receives either pure noise or corrupted NC-OFDM/OFDM signals, and builds up a dataset out of the received signals. Then, it performs a two-stage ML-based attack. First, it does spectrum sensing in order to distinguish between the two types of the received signals. Second, it performs ML algorithms on the dataset of corrupted NC-OFDM/OFDM signals aiming at spoofing the transmitter. Spectrum sensing is done via Gaussian-mixture variational autoencoder (GMVAE) \cite{GMVAE} in an unsupervised manner, i.e., it does not need labels during training. For the spoofing attack part, either supervised and unsupervised algorithms are assumed to be utilized by the adversary in order to infer PHY parameters. The former relies on true labels during training while the latter learns from the structure of raw data itself. For the supervised case, we assume that the adversary trains a deep feed-forward neural network (DNN) to estimate the transmission parameters. For the unsupervised scenario, VAEs are utilized by the adversary to extract representations from the datasets of NC-OFDM/OFDM signals that can be used for PHY spoofing. In the unsupervised scenario, we also develop a new metric based on disentanglement principles \cite{DIPVAE} to measure the usefulness of the learned representations for PHY spoofing.

We provide numerical results showing that an
adversary equipped with ML tools is able to spoof an NC-OFDM system, which
was previously considered to be completely secure against cyclostationary based
attacks. Furthermore, we demonstrate that the representations learned by the VAEs carry significant information about the PHY characteristics of the NC-OFDM signals, including the total number of subcarriers, the amount of power sent in each subcarrier and modulation schemes (e.g., BPSK and QAM), all of which can be used by the adversary for PHY spoofing. Hence, unlike what is suggested by cyclostationary analysis in \cite{RatenshMilCom,OURMILCOM}, these results show that NC-OFDM systems are vulnerable to PHY spoofing if the adversary is equipped with ML tools. However, we also establish that the success of spoofing attacks highly depends
on the subcarrier occupancy patterns chosen at the transmitter; the
more structured is the band allocation, the better is the performance
of the adversary in estimating the transmission parameters. Therefore, in order to secure the NC-OFDM systems against ML-based attacks, the transmitter should employ random subcarrier occupancy patterns, where active subcarriers are chosen in a (pseudo)random fashion to preclude the adversary from correctly inferring the transmission parameters and
spoofing the signal.

\subsection{Notation and organization}
Throughout the paper, vectors are denoted with lowercase bold letters while uppercase bold letters are reserved for matrices. The $m$th element of a vector $\mathbf{u}$ is denoted by $\mathbf{u}(m)$. Non-bold letters are used
to denote scalar values and calligraphic letters denote sets. The spaces of real and complex vectors of length $d$ are denoted by $\mathbb{R}^d$ and $\mathbb{C}^d$, respectively. Also, real and imaginary parts of a complex number $a$ are denoted by $\mathcal{R}(a)$ and $\mathcal{I}(a)$, respectively. The expectation and probability mass (or density) function
of a random variable $\mathbf w$ are denoted by $\mathbb E_{p(\mathbf w)}(\cdot)$ and $p(\mathbf w)$,
respectively, while $\Pr(\cdot)$ is used to denote the probability of an
event. 

The rest of the paper is organized as follows. The system model is described in Section \ref{sec:SystemModel}. An introduction to cyclostationary
analysis and its limitations for PHY spoofing are
presented in Section \ref{sec:Signal exploitation using cyclostationary analysis}. In Sections \ref{sec:Unsupervised learning for PHY spoofing} and \ref{sec:Supervised Learning for PHY Spoofing}, PHY spoofing attacks based on unsupervised and supervised learning algorithms are discussed, respectively. Section \ref{sec:Unsupervised learning for PHY spoofing} also describes the significance of the learned representations in VAEs for NC-OFDM/OFDM signals. We introduce a new metric based on the idea of disentanglement in Section \ref{Sec:Disentanglement metrics} to measure the usefulness of the learned representations for PHY spoofing. Finally, we present numerical examples that highlight the performance of learning-based PHY spoofing attacks in Section \ref{Sec:Numerical examples}.

\section{System model}\label{sec:SystemModel}
Consider a system composed of a transmitter (Tx), a receiver (Rx) and an adversary. An important type of attack in this setting is PHY spoofing where an adversary disguises itself as a legitimate transmitter and sends spurious data to the receiver (Fig. \ref{fig:model}). Specifically, the adversary overhears the signals sent by the Tx to the Rx, and its goal is to send bogus data to the Rx using signals that have similar PHY characteristics to the ones sent by the Tx. In this way, the Rx cannot distinguish the source of the original and the bogus data, and by decoding the latter, it might compromise the underlying system security in different ways.

The point-to-point communication link between the Tx and Rx is assumed to operate over a total bandwidth $B$ composed of a set of $N$ subcarriers. The transmitter can either transmit over the whole band in the case of OFDM transmissions, or a subset of subcarriers (known as active subcarriers) in the case of NC-OFDM transmissions. The signal transmitted by the Tx can be written as:
\begin{equation}
\label{eq:NCOFDMSignal}
s(t)=\sum_{m=-\infty}^{m=+\infty}\sum_{n=1}^N \mathbf{u}_m(n) s_{m,n}p_{m,n}e^{j2\pi f_n(t-mT_o)}g(t-mT_o),
\end{equation}
where $s_{m,n}\in\mathbb{C}$ and $p_{m,n}\in\mathbb{R}$ are the transmitted symbol and a power factor corresponding to the $n$th subcarrier in the $m$th time slot, respectively. The total duration of one NC-OFDM/OFDM symbol is given by $T_o=T_u+T_{cp}$, with $T_u$ and $T_{cp}$ being the NC-OFDM symbol duration and the duration of the cyclic prefix, respectively. Furthermore, $\mathbf{u}$ is called \textit{subcarrier occupancy pattern}, which is a binary vector of size $N$ whose elements are zero for inactive subcarriers and one for active subcarriers. Particularly, for an OFDM transmission $\mathbf{u}$ amounts to an all-one vector of size $N$. We assume $g(\cdot)$ is a rectangular pulse of width $T_o$ centered at $T_o/2$.
The center frequency of each subcarrier is denoted by $f_n=n\Delta f$ where $\Delta f=1/T_u$ represents the width of each subcarrier.

For each OFDM/NC-OFDM symbol, the Tx chooses the parameters $\Delta f$, $\mathbf{u}$ and $N$ and transmits a signal to the Rx. The positions of active subcarriers in $\mathbf{u}$ either follow a certain pattern or are totally random. The adversary seeks to find these transmission parameters in order to generate waveforms similar to (\ref{eq:NCOFDMSignal}), inject bogus data in place of $s_{m,n}$, and transmit them to the receiver. We assume the Rx only decodes data that are being sent over the active subcarriers with the correct $\Delta f$ and $N$ chosen by Tx; otherwise, a decoding failure will occur. Therefore, we utilize bit error rate (BER) at Rx as a measure to evaluate the performance of the adversary in terms of spoofing. If this BER is close to that of the baseline transmission (where the parameters are perfectly known at the Rx), it is indicative of the maximum spoofing performance of the adversary. At the other extreme, a BER close to $0.5$ suggests that the adversary cannot do much in terms of spoofing, i.e., Tx-Rx transmission is secured against PHY spoofing. We note that as Tx may use a different set of transmission parameters to transmit each NC-OFDM/OFDM symbol (e.g. sending over a different set of subcarriers), these parameters need to be estimated by the Rx as well in order to ensure reliable communication for the legitimate parties.

\subsection{Channel model}
\label{sec:Channel Model}
\begin{figure}
   \centering
\includegraphics[scale=0.32]{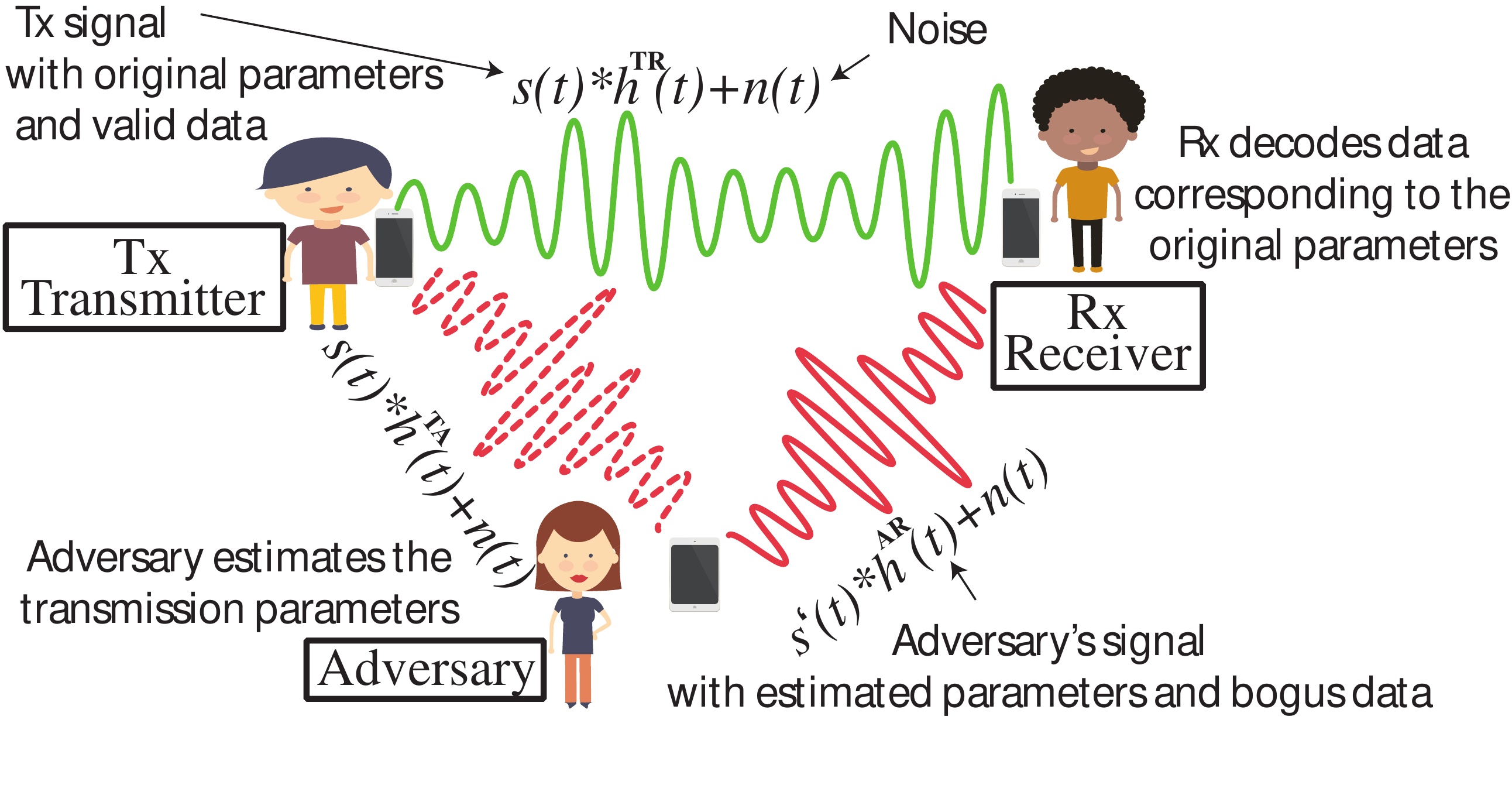}

 \caption{ PHY spoofing by an adversary overhearing Tx-Rx transmissions.}%
 \label{fig:model}
  \end{figure}

The system model in Fig. \ref{fig:model} consists of three channels corresponding to the pairs Tx-Rx (TR), Tx-Adversary (TA) and Adversary-Rx (AR), where each could have a specific channel impulse response (CIR) denoted by $h^{TR}(t)$, $h^{TA}(t)$ and $h^{AR}(t)$, respectively, and a different signal-to-noise ratio (SNR). Particularly, the spoofing performance highly depends on the SNR of the Tx-Adversary channel, which we call spoofing SNR in the rest of the paper. For all three channels, the received signal by a party is given by the convolution $r(t)=s(t)^*h(t)+n(t)$, where $n(t)$ is additive white Gaussian noise, and $h(t)$ is the corresponding CIR between the two parties. The discrete received samples are given by $r(t)=r(t_i)$, where $i=0,\dots,n_1-1$ and $n_1$ represents the number of (complex) samples. We assume noise samples at different time instances $t_i$'s are independent and identically distributed (i.i.d.) with zero mean and variance $N_0/2$. Denoting $\mathbf s=[ s(t_0),\dots,s(t_{n_1-1})]$, signal power is computed by $E_s=\lVert \mathbf s \rVert^2/n$ where $\lVert\cdot\rVert$ is the $l_2$-norm. Then, SNR and SNR per bit equal $E_s/N_0$ and  $E_b/N_0=\frac{E_s}{QN_0}$, respectively. One can verify that $Q$ equals $N_ab/N$ where $N_a$ denotes the number of active subcarriers, and $b$ is the number of bits sent over each subcarrier.

 \begin{table}[t]
 \centering
  \caption{Three sets of transmission parameters used for NC-OFDM signals in Example \ref{CAFexample}.}%
\label{CAFTable}
\begin{tabular}{|l|l|l|l|}
\hline
Case & $T_u (\mu s)$   & $T_0(\mu s)$ & $q$ \\ \hline
$1$ & $320$ & $320\leq T_0 \leq 640$  & $5$ \\ \hline
$2$ & $256$ &  $320\leq T_0 \leq 512$ & $4$ \\ \hline
$3$ & $192$ &  $320\leq T_0 \leq 384$ & $3$ \\ \hline
\end{tabular}
    \end{table}

\subsection{Adversary's model}
From the attacker's perspective, we assume the adversary has access to radio equipment for overhearing the transmissions between two legitimate parties, and is able to sample each received signal, and build up a dataset out of these samples. As the adversary is constantly listening to the transmitting terminal, these received signals may be only noise or an NC-OFDM/OFDM signal plus noise. Furthermore, we assume adversary has resources for data processing via ML algorithms, and is equipped with a radio transmitter for generating signals with desired parameters and sending them through the AR channel. Specifically, the adversary utilizes ML to perform a two-stage attack. First, it performs spectrum sensing in order to distinguish pure noise from corrupted NC-OFDM/OFDM signals. Then, a spoofing attack is carried out using only the corrupted signals. We note that each entry of the dataset may or may not be associated with the corresponding true transmission parameters ($\Delta f$, $N$, $\mathbf{u}$) referred to as labels. Depending on the availability of the labels during the training stage, two types of ML algorithms are useful: supervised and unsupervised. The former makes use of the labels for training the DNNs while the latter exploits possible data structure and clustering methods without using labels. 
\section{PHY spoofing via cyclostationary analysis}\label{sec:Signal exploitation using cyclostationary analysis}
\begin{figure}
    \centering

\includegraphics[width=8cm]{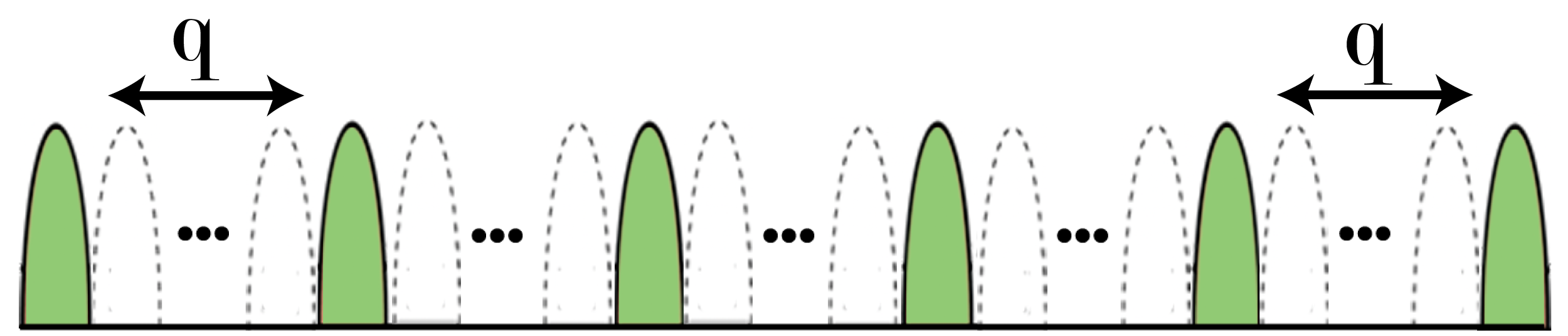}
\caption{Interleaved subcarrier occupancy pattern with interleaving factor $q$ in an NC-OFDM symbol.}
\label{fig:structNC0}

    \end{figure}

\begin{figure}
\centering
\includegraphics[width=7.5cm]{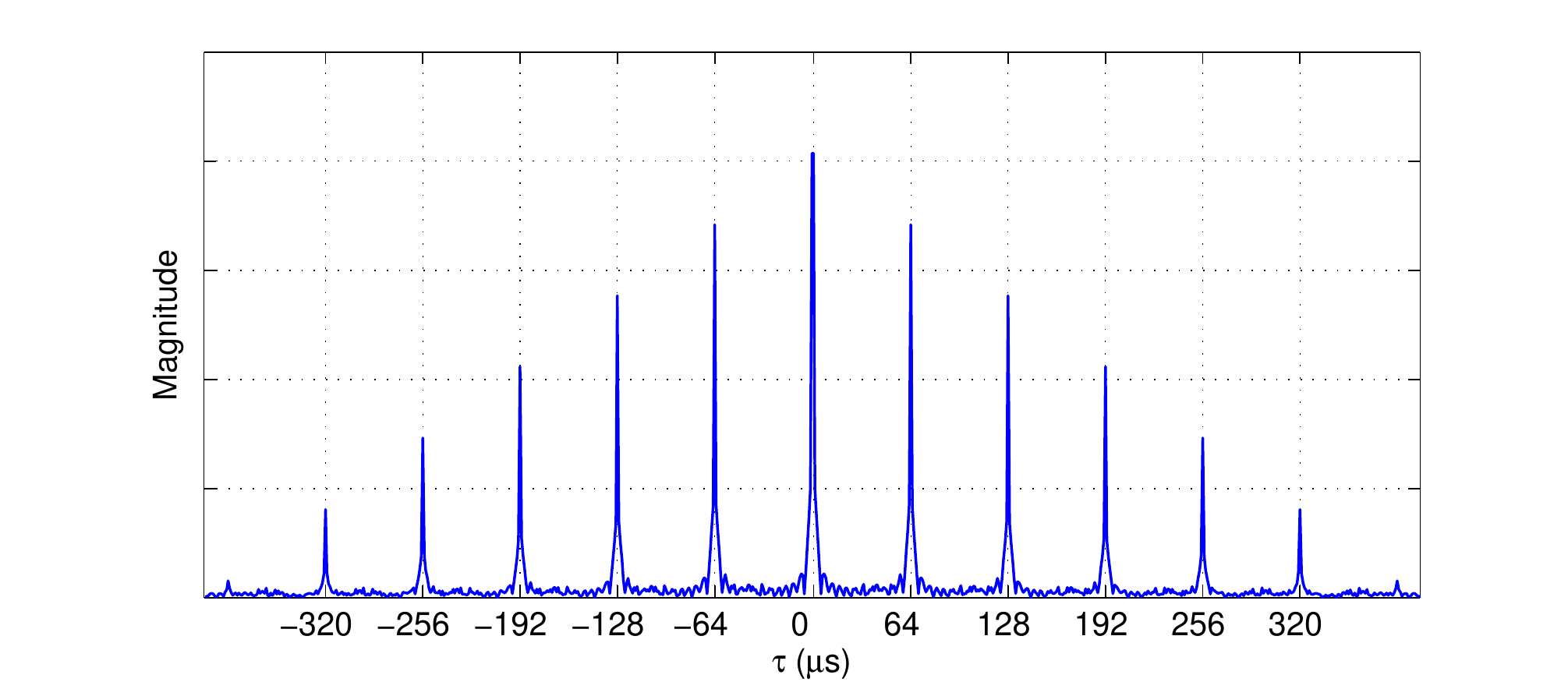}
\caption{Estimated CAF (\ref{estimatedCAF}) at $\alpha=0$ for the cases in Table \ref{CAFTable} at spoofing SNR of $5$ dB (no fading), where $\tau=\bar{\tau}T_s$, $\bar{\tau}\in[-400,400]$ and $T_s=10^{-6}.$}
\label{fig:CAF}

\end{figure}

While cyclic prefix is useful to mitigate the effect of inter-carrier interference in multicarrier systems, it also enables an adversary to infer basic transmission parameters using cyclostationary analysis \cite{CAF1}. Cyclostationary analysis is based on the auto-correlation function $R(t,\tau)$ of the transmitted signal, which is calculated as
\begin{align}
\label{autocorrelation}
\begin{split}
R(t,\tau)=E[s(t)s^*(t-\tau)]=\sigma^2_s\big(\sum_{n=1}^N\mathbf{u}(n)e^{j2\pi f_n\tau}\big)\\
\sum_{m=-\infty}^{m=\infty} g(t-mT_0)g^*(t-mT_0-\tau),
\end{split}
\end{align}
where $\sigma^2_s=\mathbb{E}[|s_{m,n}|^2]$. The periodicity of $R(t,\tau)$ in $t$ allows representing it as a Fourier series sum
 \begin{align}
\label{autocorrelation2}
\begin{split}
R(t,\tau)&=\sum_{n=1}^N \mathbf{u}(n)R(\alpha_n,\tau)e^{j2\pi\alpha_nt},
\end{split}
\end{align}
where $\alpha_n=n/T_0$ is the cyclic frequency and $R(\alpha_n,\tau)$ is called the cyclic auto-correlation function (CAF). For OFDM transmissions, this function can provide an adversary with $\Delta f=1/T_u$ \cite{CAF1,CAF2}. However, for NC-OFDM transmissions this analysis does not always lead to the correct results as illustrated in the following example.
\begin{example}
\label{CAFexample}
Consider an NC-OFDM signal with a total number of subcarriers $N=64$ with occupancy pattern vector $\mathbf{\mathbf{u}}$ illustrated in Fig. \ref{fig:structNC0}, where active subcarriers are spaced $q$ subcarriers apart (known as interleaved subcarriers). The transmitter chooses two transmission parameters $q$ and $T_u$ based on one of the cases listed in Table \ref{CAFTable} and sends (\ref{eq:NCOFDMSignal}) over the channel. Then, the adversary receives the noisy signal, samples it at an arbitrary rate $1/T_s$ above Nyquist frequency to obtain $M$ samples and estimates the CAF function using
\begin{equation}
\label{estimatedCAF}
\hat{R}(\alpha,\bar{\tau}T_s)=\frac{1}{M}\sum_{n=1}^Mr[n]r^*[n-\bar{\tau}]e^{-j2\pi \alpha n T_s},
\end{equation}
where $r[n]$ denotes the $n$th obtained sample and $\bar{\tau}$ belongs to the set of integers. To extract $q$ and $T_u$ corresponding to the original transmission, the adversary must look at the locations of the absolute peaks in (\ref{estimatedCAF}) at $\alpha=0$ as illustrated in Fig. \ref{fig:CAF}. We note that for all three cases, CAF-based analysis results in the same plot. In other words, the adversary cannot decide which set of transmission parameters is used by the transmitter.
 \hfill $\blacksquare$
\end{example}


In the next sections, we study how the adversary can make use of deep learning models for PHY spoofing. Our main motivation for taking this approach comes from the fact that analytical approaches seem to be failing at spoofing even for simple NC-OFDM systems, which might give PHY designers the idea that the NC-OFDM PHY is secure. We investigate the PHY spoofing performance of an adversary equipped with deep learning tools, which of course comes at the expense of a higher cost (associated with training), and answer the question that if/when it is able to do so and what are the parameters that affect its performance. 

\section{PHY spoofing via unsupervised learning}
\label{sec:Unsupervised learning for PHY spoofing}
\begin{figure}

        \centering
        \includegraphics[width=8.8cm]{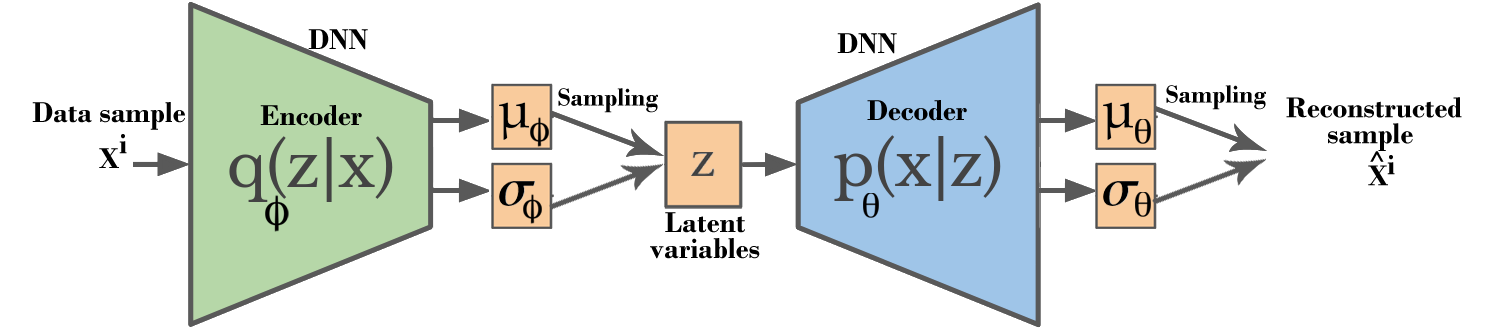}
        \caption{A schematic of a VAE where encoder and decoder are chosen to be Gaussian distributions with mean vector $\mathbf{\mu}$ and diagonal covariance matrix  with variances $\mathbf{\sigma}^2$. These are parameterized by two DNNs trained based on (\ref{eq:VAEobjectivefunc}). $\mathbf{z}$ and $\hat{\mathbf{x}}^i$ are generated by sampling from the distributions corresponding to the encoder and the decoder, respectively. }
        \label{Fig:VAE}
  \end{figure}

We assume the adversary utilizes variational autoencoders (VAEs) for PHY spoofing in an unsupervised manner. VAEs are designed based on the idea of variational inference, which can be described via a latent variable model (LVM). An LVM is a generative model for a dataset $\mathbf{\mathcal{D}}=\{\mathbf{x}^i\}_{i=1}^{M}$ consisting of $M$ i.i.d. samples of a continuous random variable $\mathbf{x}$. LVM is defined over a joint distribution $p_{\mathbf{\theta}}(\mathbf x,\mathbf z)$ parameterized by $\mathbf{\theta}\in\mathbb{R}^j$ where $\mathbf z$ is an unobserved continuous random variable known as the latent variable (feature space) and $j$ denotes the model size. The joint density $p_{\theta}(\mathbf x,\mathbf z)$ is denoted by $p_{\theta}(\mathbf x,\mathbf z)=p(\mathbf z)p_{\theta}(\mathbf x|\mathbf z)$ where $p(\mathbf z)$ is a fixed prior over the latent space and $p_{\theta}(\mathbf x|\mathbf z)$ is the conditional generator. Then, the inference problem amounts to finding the posterior of the hidden random variable $\mathbf z$, i.e., $p_{\theta}(\mathbf z|\mathbf x)=\frac{p_{\theta}(\mathbf x,\mathbf z)}{\int p_{\theta}(\mathbf x,\mathbf z)d\mathbf z}$. However, this integration is not tractable for complicated datasets $\mathbf{\mathcal{D}}$ consisting of high-dimensional data samples $\mathbf{x}^i$'s, and alternative methods must be utilized in practice to compute it approximately. Variational inference is a promising optimization-based candidate for this purpose, and has enjoyed a lot of attention during the past few years \cite{VAE,DIPVAE,FactorVAE}. Particularly, a VAE learns an optimal model ($\mathbf{\theta}^*$) which maximizes the probability of data samples by maximizing the expectation of the evidence probabilities, i.e.,
\begin{equation}
\label{eq:expectedmarginal}
    \mathbb{E}_{p_{}(\mathbf x)}[\log p_{\theta}(\mathbf x)]= \mathbb{E}_{p_{}(\mathbf x)}[\log \mathbb{E}_{p(\mathbf z)}[p_{\theta}(\mathbf x|\mathbf z)]],
\end{equation}
over $\mathbf{\theta}$ where $p(\mathbf x)$ denotes the underlying true distribution of the data in the dataset $\mathbf{\mathcal{D}}$. However, this requires computing $\int_{\mathbf z} p_{\theta}(\mathbf x|\mathbf z)p(\mathbf z)d\mathbf z$ which is not tractable. Instead, \cite{VAE} obtains a lowerbound called evidence lower bound (ELBO) for individual data sample $\mathbf{x}^i$ in (\ref{eq:expectedmarginal}), which is given by
\small
\begin{equation}
\begin{split}
\label{eq:ELBO}
\mathbf{\mathcal L}_{ELBO}(\mathbf x^i)=-\mathbb D_{KL}(q_{\phi}(\mathbf z|\mathbf x^i)||p(\mathbf z))+\mathbb E_{q_{\phi}(\mathbf z|\mathbf x^i)}[\log p_{\theta}(\mathbf x^i|\mathbf z)]
\end{split}.
\end{equation}
\normalsize
 Then, a VAE solves the following optimization problem,
 \begin{equation}
 \label{eq:VAEobjectivefunc}
     \max_{\mathbf{\theta},\mathbf{\phi}}\ \  \mathbb{E}_{p_{}(\mathbf x)}[\mathbf{\mathcal L}_{ELBO}(\mathbf x)]
 \end{equation}
 where $\mathbf{\theta}\in\mathbb{R}^j$ and $\mathbf{\phi}\in\mathbb{R}^h$ are optimization variables of dimension $j$ and $h$, respectively. KL divergence between two continuous random variables with distributions $p(x)$ and $q(x)$ is defined as $\mathbb D_{KL}(p||q)=\int_{-\infty}^{\infty}\ p(x)\log(\frac{p(x)}{q(x)})dx$. We call $q_{\phi}(\mathbf z|\mathbf x^i)$ a probabilistic encoder, since given a data sample $\mathbf{x}^i$ it produces a distribution (e.g. a Gaussian)
over the possible values of the code $\mathbf{z}$ from which the data sample $\mathbf{x}^i$ could have been generated. Similarly, we call $p_{\theta}(\mathbf x^i|\mathbf z)$ a probabilistic decoder. We choose both encoder and decoder to be Gaussian with diagonal covariance matrix whose parameters are estimated by DNNs $\mathbf{\theta}$ and $\mathbf{\phi}$, respectively (see Fig. \ref{Fig:VAE}).

The $\mathbb D_{KL}$ (KL divergence) term in (\ref{eq:ELBO}) can be seen as a regularizer that encourages the posterior $q_{\mathbf{\phi}}(\mathbf z|\mathbf x)$ to be close to the prior $p(\mathbf z)$, and the second term is called reconstruction loss. Particularly, when ELBO is maximized the KL-divergence term approaches zero. This KL term can be computed analytically as a closed-form expression by choosing specific distributions, e.g., Gaussian and Bernoulli distributions, for $q_{\phi}(\mathbf z|\mathbf x)$ and $p(\mathbf z)$ \cite{VAE}. Afterward, one can efficiently minimize the negative of ELBO using the mini-batch gradient descent algorithm \cite{VAE}. Mini-batch gradient descent is a variant of the gradient descent algorithm that splits the training dataset into small batches that are then used to calculate the loss function followed by updating the model parameters.

One of the long-standing problems in ML literature is learning representations from large datasets in an unsupervised manner that facilitates the downstream learning tasks (e.g., classification). VAEs have shown great potential \cite{BetaVAE} for learning such so-called ``useful" representations. Although it is not clear how to define/measure usefulness in unsupervised settings, researchers have developed the concept of disentangled representation that offers several advantages. Data samples in this concept are assumed to be generated via a finite number of generative factors $k_1,\dots,k_f$, each representing notions like position, scale, rotation, etc., in the case of an image sample for instance. Although there is no canonical definition for a disentangled representation, \cite{FactorVAE} points out that the learned representations are called disentangled if changes in one of the generative factors of the data are mirrored by the encoder in Fig. \ref{Fig:VAE} in exactly one of the latent variables (i.e., one dimension of the latent space). We will discuss this in more depth in Section \ref{Sec:Disentanglement metrics}, mention the weaknesses associated with this definition, and propose a new metric to overcome some of them for the datasets of NC-OFDM/OFDM signals. A VAE can encourage learning disentangled representations by choosing $p(\mathbf z)\sim\mathcal{N}(\mathbf 0,\mathbf I)$, which corresponds to different latent variables being uncorrelated. However, this choice is not enough to achieve a disentangled representation from large datasets with high-dimensional data samples since the VAE objective function in (\ref{eq:ELBO}) usually incurs a trade-off between the reconstruction loss and the KL divergence term that determines the disentanglement level. Hence, researchers have been looking for ways to not sacrifice one for the other by redesigning VAEs. In this paper, we will focus on four of the most effective techniques for learning disentangled representations and investigate their performances in the context of spoofing NC-OFDM/OFDM signals.
\begin{itemize}
\item \textit{$\beta$-VAE}:
The authors in \cite{BetaVAE} proposed to weight the KL divergence term in (\ref{eq:ELBO}) by a real-valued factor $\beta>1$ to encourage the VAE to learn disentangled representations. Therefore, the ELBO for data sample $\mathbf{x}^i$ in $\beta$-VAE is
\begin{align}
\label{eq:BetaELBO}
\mathbf{\mathcal L}^{\beta}_{ELBO}(\mathbf{x}^i)=&-\beta\
\mathbb D_{KL}(q_{\phi}(\mathbf z|\mathbf{x}^i)||p(\mathbf z))\nonumber\\&+\mathbb E_{q_{\phi}(\mathbf z|\mathbf{x}^i)}[\log p_{\theta}(\mathbf{x}^i|\mathbf z)].
\end{align}
An ELBO with a larger $\beta$ encourages learning disentangled representations as it penalizes more the dissimilarities between the learned posterior and the prior $p(\mathbf{z})\sim\mathcal{N}(\mathbf 0,\mathbf I)$ by weighing the KL term. However, this has been shown to sacrifice the reconstruction ability of the VAE in large datasets \cite{BetaVAE}.
\item \textit{DIP-VAE}:
Disentangled inferred prior (DIP) VAE was proposed in \cite{DIPVAE}. The authors add an extra term $\mathbb{D}_{KL}(q_{\phi}(\mathbf z)||p(\mathbf z))$ to ELBO (\ref{eq:ELBO}) where $q_{\phi}(\mathbf z)=\int q_{\phi}(\mathbf z|\mathbf x)p(\mathbf x)d\mathbf x$ is the learned prior. By explicitly minimizing this term one can effectively encourage learning disentangled latent variables without the need for weighting the $KL$ term in ELBO, which could result in higher reconstruction losses. In this vein, DIP-VAE matches the moments of two distributions $q_{\phi}(\mathbf z)$ and $p(\mathbf z)$. In particular, it encourages the covariance of $q_{\phi}(\mathbf z)$ to be the same as the covariance of $p(\mathbf z)$ to minimize $\mathbb{D}_{KL}(q_{\phi}(\mathbf z)||p(\mathbf z))$. Using the law of total covariance, the covariance of $\mathbf z\sim q_{\phi}(\mathbf z)$ can be written as
\small

\begin{align}
\label{eq:DIPVAE1}
    Cov_{q_{\phi}(\mathbf z)}(\mathbf z):=&\mathbb{E}_{q_{\phi}(\mathbf{z})}\big[(\mathbf{z}-\mathbb{E}_{q_{\phi}(\mathbf{z})}[\mathbf{z})](\mathbf{z}-\mathbb{E}_{q_{\phi}[\mathbf{z}]}[\mathbf{z}])^T\big]\nonumber\\=&\mathbb{E}_{p(\mathbf x)}[Cov_{q_{\phi}(\mathbf z|\mathbf{x})}(\mathbf z)]+Cov_{p(\mathbf{x})}[\mathbb{E}_{q_{\phi}(\mathbf z|\mathbf{x})}(\mathbf z)].
    \end{align}

\normalsize
For a dataset with real-valued data samples, $q_{\phi}(\mathbf z|\mathbf{x})$ can be chosen to be $\mathcal{N}\big(\mu_\phi(\mathbf x),\Sigma_\phi(\mathbf x)\big)$ \cite{VAE} where $\mu_\phi(\mathbf x)$ and $\Sigma_\phi(\mathbf x)$ represent mean and covariance of the Gaussian $\mathbf{z}$ for a given $\mathbf{x}$ as a function of the parameter $\mathbf{\phi}$. Then, plugging the mean and covariance of $q_{\phi}(\mathbf z|\mathbf{x})$ in (\ref{eq:DIPVAE1}), we get $Cov_{q_{\phi}[\mathbf z]}(\mathbf z)=\mathbb E_{p(\mathbf x)}[\Sigma_\phi(\mathbf x)]+Cov_{p(\mathbf{x})}[\mu_\phi(\mathbf x)]$ which has to be close to the identity matrix for the case when $p(\mathbf z)\sim \mathcal{N}(\mathbf 0, \mathbf I)$. By using $l_2$-norm as the measure of the proximity, DIP-VAE solves the following optimization problem
\small
\begin{equation}
\begin{aligned}
\label{eq:DIPVAE2}
    \max_{\theta,\phi}\ \  &\mathbb{E}_{p(\mathbf{x})}\big[\mathcal{L}_{ELBO}(\mathbf x)\big]- \lambda_{od}\sum_{i\neq j}[Cov_{p(\mathbf{x})}[\mu_\phi(\mathbf x)]]^2_{ij}- \\&\lambda_d\sum_i([Cov_{p(\mathbf{x})}[\mu_\phi(\mathbf x)]]_{ii}-1)^2,\end{aligned}
\end{equation}
\normalsize
where $\mathcal{L}_{ELBO}$ is the same as (\ref{eq:ELBO}), and $\lambda_{d}$ and $\lambda_{od}$ are two hyper-parameters controlling penalties induced by the diagonal and the off-diagonal components in $Cov_{q_{\phi}(\mathbf z)}(\mathbf z)$, respectively. 

\item \textit{FactorVAE}:
As mentioned previously, weighting the KL term $\mathbb D_{KL}(q_{\phi}(\mathbf z|\mathbf x)||p(\mathbf z))$ by a $\beta>1$ could negatively impact the reconstruction performance of VAE. Mathematically, this can be seen from
\small
\begin{equation}
\label{eq:FactorVAE1}
    \mathbb{E}_{p(\mathbf x)}[\mathbb{D}_{KL}(q_{\phi}(\mathbf z|\mathbf x)||p(\mathbf z))]=\mathbb I(\mathbf x;\mathbf z)+\mathbb{D}_{KL}(q_{\phi}(\mathbf z)||p(\mathbf z)),
\end{equation}
\normalsize
where $\mathbb I(\mathbf x;\mathbf z)$ is the mutual information between $\mathbf x$ and $\mathbf z$. Therefore, penalizing $\mathbb{D}_{KL}(q_{\phi}(\mathbf z|\mathbf x)||p(\mathbf z)$ is a double-edged sword. On the one hand, it forces $q_{\phi}(\mathbf z)$ to be close to $p(\mathbf z)=\mathcal{N}(\mathbf 0,\mathbf I)$, and encourages learning disentangled representations. On the other hand, by penalizing $\mathbb I(\mathbf x;\mathbf z)$ it encourages learning a $\mathbf z$ independent of $\mathbf x$, which would limit the amount of information stored in $\mathbf{z}$ about $\mathbf{x}$. Thus a larger $\beta$ leads to a better disentanglement but reduces the reconstruction quality. Similar to DIP-VAE, FactorVAE avoids this conflict by augmenting $\mathbb{D}_{KL}(q_{\phi}(\mathbf z)||p(\mathbf z))$ to the ELBO (\ref{eq:ELBO}) in order to directly encourage independence in the latent variables, which results in the following objective:
\begin{equation}
\begin{aligned}
\label{eq:FactorVAE}
\max_{\theta,\phi}&\ \mathbb{E}_{p(\mathbf x)}\big[\mathbb E_{q_{\phi}(\mathbf z|\mathbf x)}[\log p_{\theta}(\mathbf x|\mathbf z)]\big]-\\&\mathbb{E}_{p(\mathbf x)}\big[\mathbb D_{KL}(q_{\phi}(\mathbf z|\mathbf x)||p(\mathbf z))\big]-\gamma\mathbb{D}_{KL}(q_{\phi}(\mathbf z)||p(\mathbf z)),\end{aligned}
\end{equation}
where $p(\mathbf z)$ is assumed to be of the form $\prod_{j=1}^{J} p(z_j)$ and $J$ denotes dimension of $\mathbf{z}$. Although the idea of directly minimizing $\mathbb{D}_{KL}(q_{\phi}(\mathbf z)||p(\mathbf z))$ was used in DIP-VAE as well, FactorVAE takes a different approach towards achieving this goal. Specifically, FactorVAE estimates the density ratio,
\begin{equation}
\label{eq:FactorVAE3}
    \mathbb{D}_{KL}(q_{\phi}(\mathbf z)||p(\mathbf z))=\mathbb{E}_{q_{\phi}(\mathbf z)}\big[\log\frac{q_{\phi}(\mathbf z)}{p(\mathbf z)}\big],
\end{equation}
 via density-ratio trick \cite{DensityRatio}. For each mini-batch, it generates samples from both $q_{\phi}(\mathbf z)$ and $p(\mathbf z)$, and approximates (\ref{eq:FactorVAE3}) by a model $D_\mathbf{\psi}$ parameterized by a DNN $\mathbf{\psi}$, which takes a sample $\mathbf{z}$ as input and outputs the probability that $\mathbf{z}$ belongs to $q_{\phi}(\mathbf z)$. Then, utilizing
\begin{equation}
\label{eq:FactorVAE4}
    \mathbb{D}_{KL}(q_{\phi}(\mathbf z)||p(\mathbf z))=\mathbb{E}_{q_{\phi}(\mathbf z)}\big[\log\frac{D_{\mathbf{\psi}}(\mathbf z)}{1-D_{\mathbf{\psi}}(\mathbf z)}\big],
\end{equation}
(\ref{eq:FactorVAE}) can be jointly maximized over the set of parameters $\mathbf{\psi},\mathbf{\theta},\mathbf{\phi}$, each of which is taken to be a DNN in this work.
\end{itemize}

\subsection{Learning useful representations from NC-OFDM signals}\label{sec:Learning useful representations from NC-OFDM signals}
In this section, we describe training of VAEs on a dataset $\mathcal{D}$ of NC-OFDM signals whose entries consist of samples of an NC-OFDM signal $s(t)$ in (\ref{eq:NCOFDMSignal}) obtained by
\begin{equation}
\label{eq:sampledNCOFDM}
s(kT_s)=\sum_{n=1}^{N} \mathbf{u}(n) s_{0,n}e^{j2\pi f_n(kT_s)}, k=0,\dots,n_1-1,
\end{equation}
where $1/T_s$ denotes the sampling rate. We concatenate the real and imaginary parts of the samples in (\ref{eq:sampledNCOFDM}) to build a single sample $\mathbf{x}^i$ of size $d=2n_1$ in the form $\mathbf{x}^i=[\mathcal R(s(0)), \mathcal R(s(T_s)), \dots, \mathcal R(s((n_1-1)T_s)), \mathcal I(s(0)), \mathcal I(s(T_s)), \dots, \mathcal I(s((n_1-1)T_s))]$ for $\mathcal{D}$. We have chosen encoder $q_\phi(\mathbf z|\mathbf x)$ and decoder $p_\theta(\mathbf x|\mathbf z)$ to be Gaussian with diagonal covariance matrix having means $\mathbf{ mu}_\phi$ and $\mathbf \mu_\theta$, and variances $\mathbf \sigma^2_\phi$, $\mathbf \sigma^2_\theta$, respectively. We note that the learned representations $\mathbf{z}$ are distributed around the mean of $q_\phi(\mathbf z|\mathbf x)$, denoted by $\mathbf{ \mu}_\phi$, and they approach $\mathbf{ \mu}_\phi$ when $\mathbf \sigma^2_\phi$ goes to zero. Encoder (resp., decoder) is modeled with a fully connected feed-forward DNN whose outputs are $\mu_\phi$ and $\mathbf \sigma^2_\phi$ (resp., $\mathbf \mu_\theta$ and $\mathbf \sigma^2_\theta$). These DNNs have $5$ hidden layers with $200$, $400$, $600$, $400$, and $200$ number of neurons in each layer. We also have trained DNNs of larger parameter spaces, but the performance of the networks did not improve noticeably. The prior $p(\mathbf z)$ is also chosen to be a Gaussian distribution with zero mean and identity covariance matrix, which makes it possible to compute the KL term in (\ref{eq:ELBO}) analytically \cite{VAE}. We build two NC-OFDM datasets whose signals are generated by two different subcarrier occupancy patterns, and investigate the properties of the learned representations by the FactorVAE (\ref{eq:FactorVAE}) with $\gamma=5$, which has shown to be able to find disentangled representations effectively (see Section \ref{Sec:Numerical examples}). Training is done via mini-batches of size $100$ with a learning rate of $0.0005$ over a dataset of size $500,000$. In order to study what information has been encoded to each dimension of the latent space, we use a common technique called \textit{latent traversal}. After training a VAE, latent traversal obtains the representation $\mathbf{z}$ corresponding to a data sample $\mathbf{x}^i$. To study what information the $i$th latent variable ($z_i$) represents about $\mathbf{x}$, latent traversal changes the value of $z_i$ (e.g. between $[-3,3]$) while fixing the other latent variables and studies the corresponding changes induced by the decoder in the fast Fourier transform (FFT) of the reconstructed sample $\hat{\mathbf{x}}$.

\begin{figure}
    \centering
        \includegraphics[width=4cm]{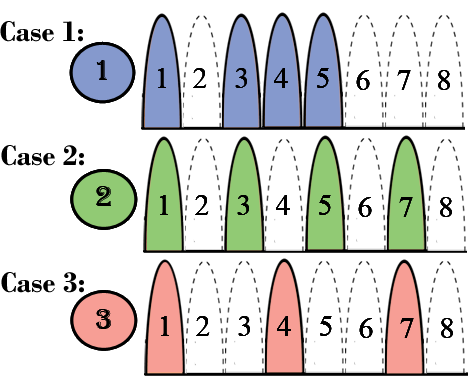}
        \caption{Three different band allocations based on which we have generated the NC-OFDM signals in the dataset. An inactive subcarrier is denoted by dashed lines.}
        \label{Fig:BandAllocations}
\end{figure}

\begin{figure}
        \centering
        \includegraphics[width=5cm]{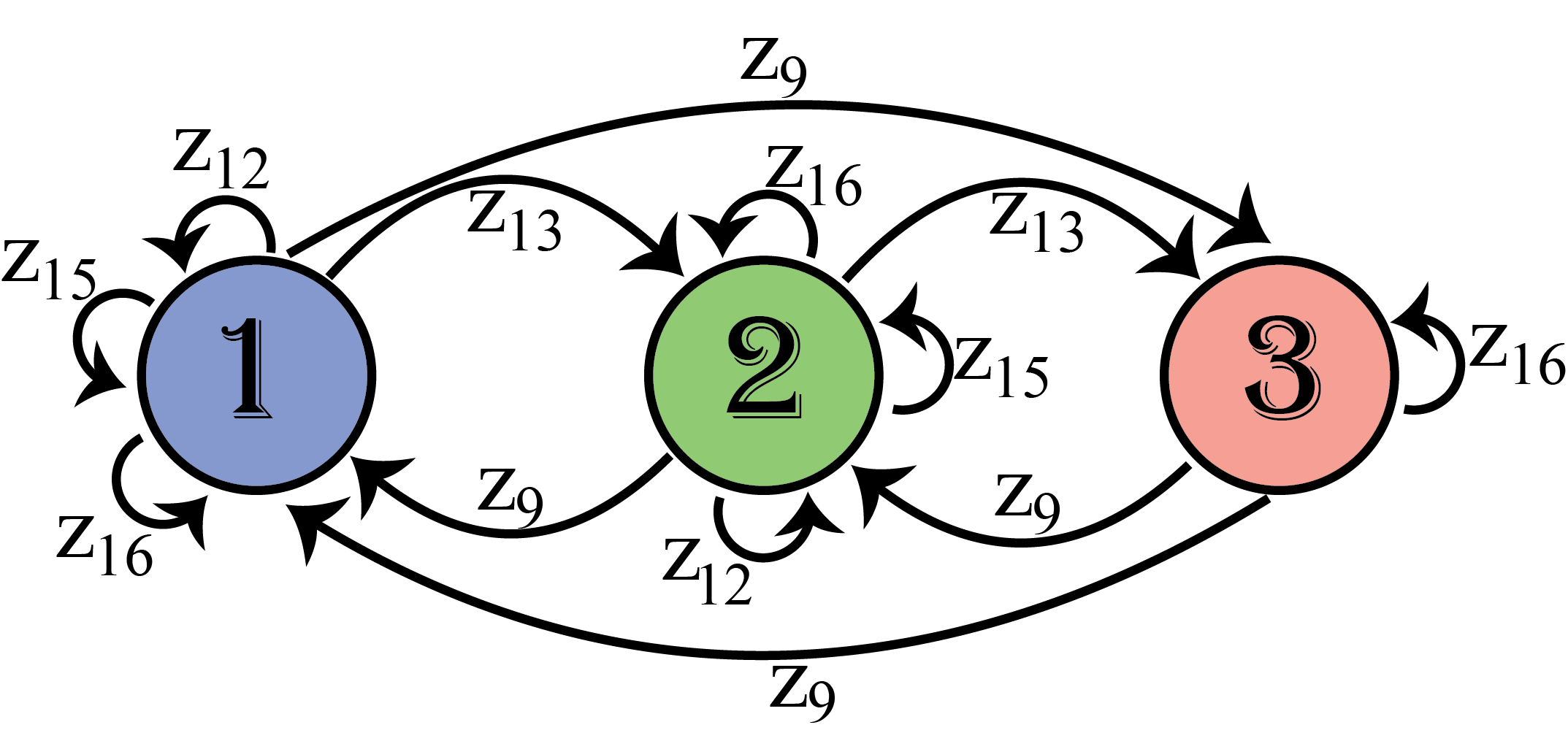}
        \caption{VAE maps the NC-OFDM signals to a continuous space $\mathbf{z}$. Latent traversal for $z_i$ is shown by an arrow where $\mathbf{x}^i$ and $\hat{\mathbf{x}}^i$ belong to the cases denoted by the arrow's tail and tip, respectively.
        }
        \label{Fig:Transitions}
\end{figure}

\begin{figure}

 \centering
 \includegraphics[width=5.5cm]{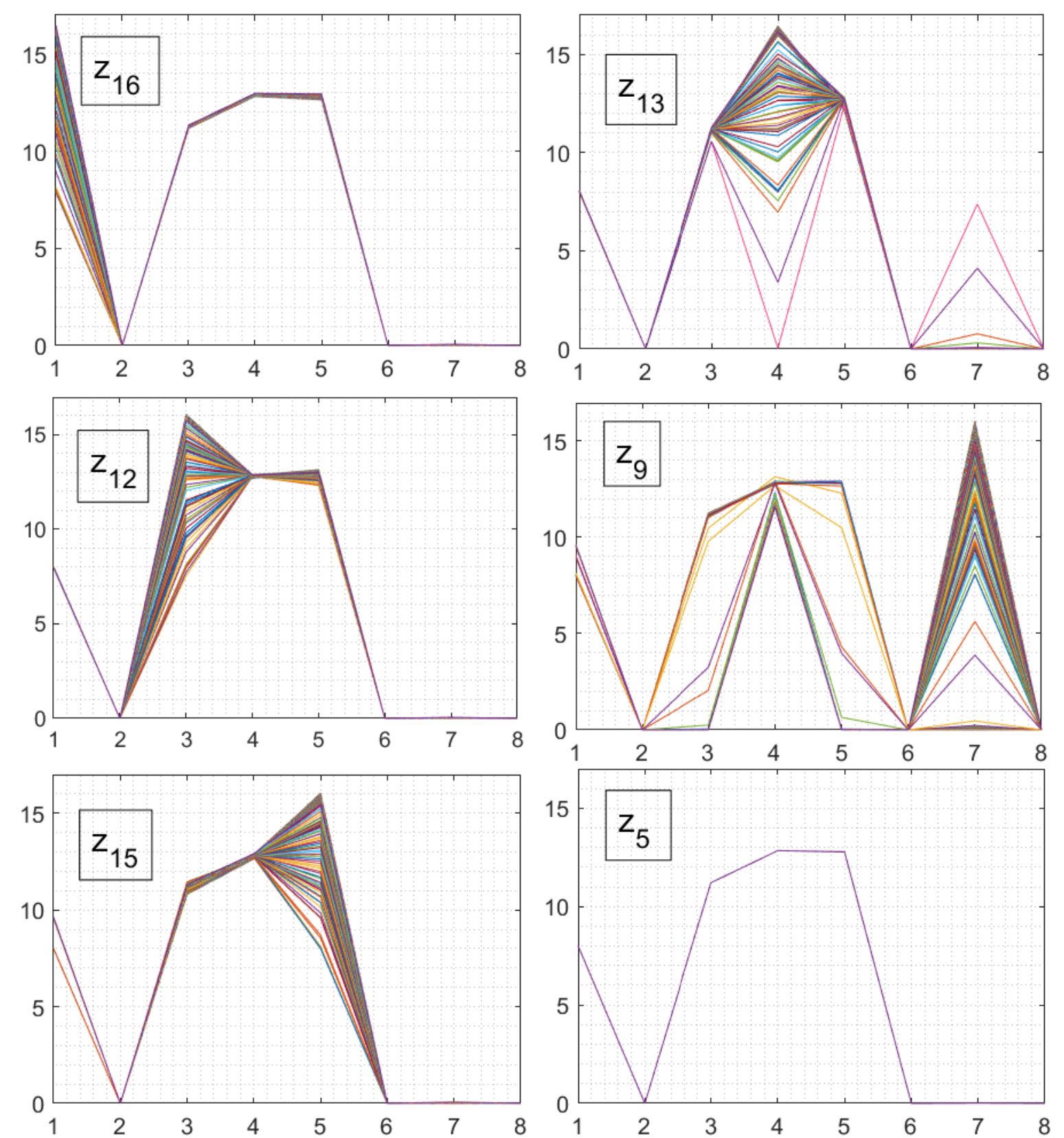}
        \caption{Latent traversal for a dataset of NC-OFDM signals based on Fig. \ref{Fig:BandAllocations}. $\mathbf{x}$ belongs to Case $1$. FFTs ($8$ points) of the reconstructed signals $\mathbf{\hat{x}}$ are depicted. }
        \label{Fig:StructuredAllocation}
   \end{figure}

First, we consider a dataset of NC-OFDM signals with a structured band allocation based on $3$ different occupancy pattern vectors depicted in Fig. \ref{Fig:BandAllocations}. We note that this is given as a toy example, which enables us to fully describe the properties of the learned latent space by a VAE through choosing a small number of subcarriers $N=8$ with only $3$ distinct occupancy patterns. The number of latent variables in FactorVAE is set to $N_{\mathbf{z}}=16$, $p_n\in [1,2]$, $n=\dots,N$, where a larger $p_n$ indicates a higher SNR for a fixed noise variance $N_0$. Binary phase-shift keying (BPSK) is utilized as the modulation technique at the transmitter. Also, $n_1=16$ complex samples are collected from each signal (\ref{eq:sampledNCOFDM}) for building the dataset. We have done latent traversal on a trained VAE for latent variables, six of which are depicted for instance in Fig. \ref{Fig:StructuredAllocation}, where the input signal $\mathbf{x}$ to the VAE is a signal from Case $1$ depicted in Fig. \ref{Fig:BandAllocations}.
We observe that the VAE only encodes information in $5$ distinct latent variables $z_{16}$, $z_{12}$, $z_{13}$, $z_{15}$ and $z_{9}$, which are called informative latent variables and control the amount of power in $5$ distinct active subcarriers depicted in Fig. \ref{Fig:BandAllocations}.
The other $11$ variables are uninformative (e.g., $z_5$) that carry no information about $\mathbf{x}$ as changing them does not have any effect on the reconstructed sample.
As we chose $p(\mathbf z)\sim\mathcal{N}(0,\mathbf I)$, we note that the learned representations lie within a continuous space whose dimensions capture the amount of power in different subcarriers as suggested by Fig. \ref{Fig:StructuredAllocation}. Specifically, changing an informative latent variable could result in generating an output $\hat{\mathbf{x}}$ by the decoder that belongs to a different case than that of $\mathbf{x}$. For example, by changing $z_{13}$ (and fixing the other variables), a signal from Case $2$ can be generated while the input signal belongs to Case $1$. One can interpret this as the decoder is changing the amount of power in subcarriers $7$ and $4$ through changing $z_{13}$. Similarly, by changing $z_9$ a signal from Case $3$ can be generated. This is illustrated in Fig. \ref{Fig:Transitions} for the whole space of $\mathbf{z}$, which shows how the VAE exploits the subcarrier occupancy pattern for finding a continuous representation space that covers all signals in the training dataset.

\begin{figure}
        \centering
        \includegraphics[width=5.5cm]{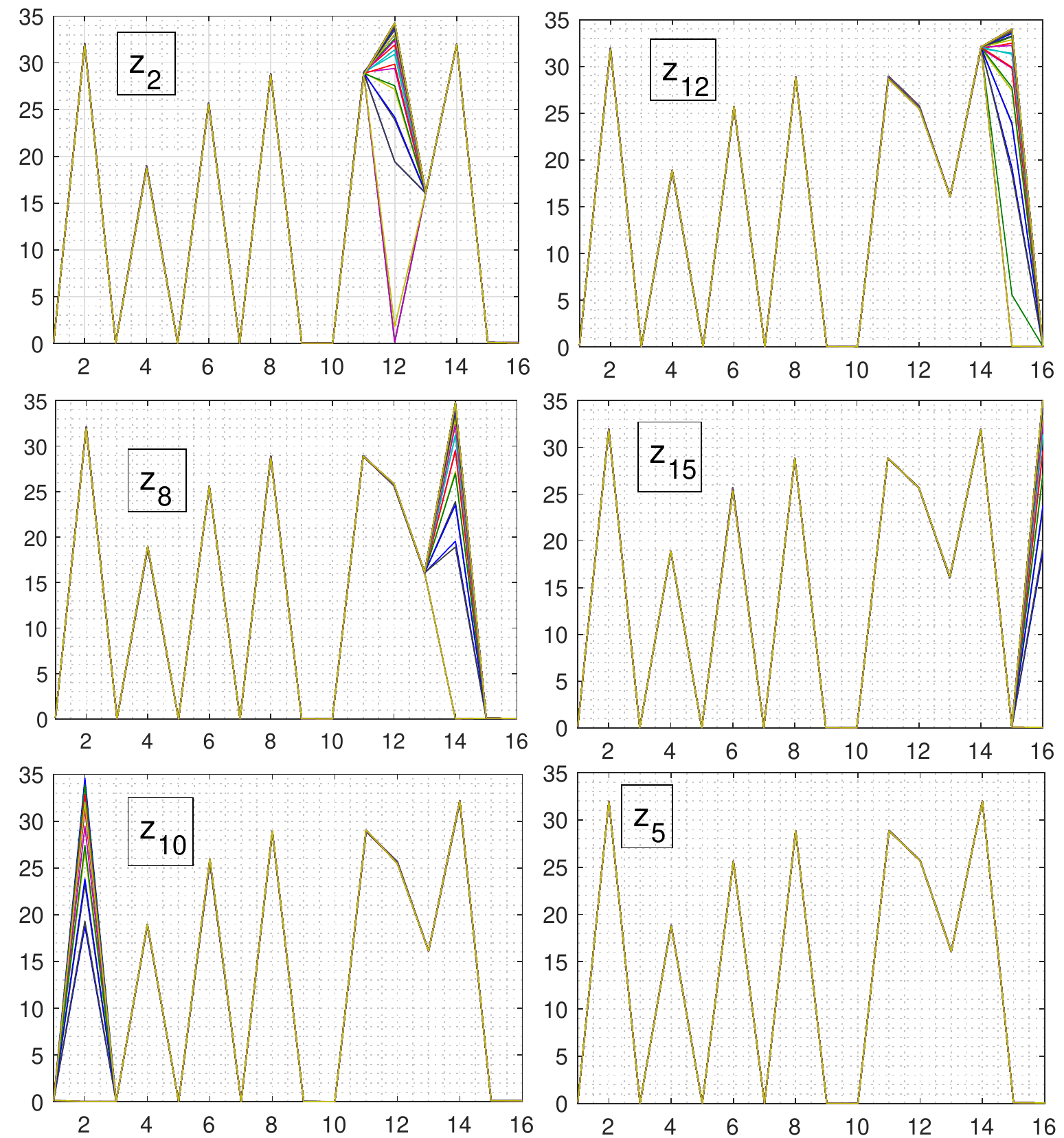}
        \caption{Latent traversal for a dataset of NC-OFDM signals with random band allocation. FFTs (16 points) of the reconstructed signals $\mathbf{\hat{x}}$ are depicted. }
        \label{Fig:RandomAllocation}

\end{figure}

Next, we consider NC-OFDM signals with random occupancy pattern vectors where active subcarriers are chosen in a random fashion, i.e., each element in $\mathbf{\mathbf{u}}$ is $0$ or $1$ with probability $0.5$. Similar to the structured case, $p_n\in[1,2]$ and BPSK is used as the modulation technique. We consider the total number of subcarriers to be $N=16$ ($2^{16}$ different subcarrier occupancy patterns), number of latent variables $N_\mathbf z=20$ and $n_1=32$. Latent traversal, in this case, shows that informative latent variables control the amount of power in exactly one subcarrier in this case (Fig. \ref{Fig:RandomAllocation}). 
As an example, one can see that changing $z_2$ results in changing the power in the $12$th subcarrier of $\mathbf{\Hat{x}}$. In other words, $z_2$ represents the relative amount of power in the $12$th subcarrier.
A similar graph to Fig. \ref{Fig:Transitions} can be sketched for this case as well where there are $2^{16}$ different band allocations. We note that $16$ out of $20$ latent variables correspond to $16$ different subcarriers, and the remaining $4$ variables are uninformative. Therefore, one can see that VAE is capable of learning a representation $\mathbf{z}$ where each latent variable $z_i$ corresponds to the relative amount of power in a unique subcarrier.

\subsection{Spectrum sensing via VAEs}\label{sec:Spectrum sensing via VAEs}
For an adversary that is constantly listening to the transmitting terminal for a certain amount of time, the received signals could either be just noise or corrupted NC-OFDM signals. PHY spoofing involves processing the latter in order to infer certain transmission parameters. Therefore, it is important for the adversary to distinguish between the noise only signals and the transmitted signals; this process is known as spectrum sensing. We utilize Gaussian mixture (GM) VAEs \cite{GMVAE} for this purpose, which similar to VAEs, are generative models built upon the idea of variational inference. However, unlike VAEs which assume the data are generated from one continuous latent variable $\mathbf{z}$, GMVAEs consider one continuous and one discrete latent variable denoted by $\mathbf{z}$ and $\mathbf{y}$, respectively, where $\mathbf{z}$ itself is also assumed to be generated from a continuous latent variable $\mathbf{w}$. ELBO for GMVAEs is obtained in \cite{GMVAE} for the case where the prior distribution of $\mathbf{z}$ and $\mathbf{w}$ is Gaussian, while $\mathbf{y}$ follows a categorical distribution with a predefined number of classes denoted by $k$. Once trained to maximize the ELBO, GMVAEs cluster the data into $k$ different groups based on their distributions. For the PHY spoofing scenario where the training dataset consists of signals belonging to either of the two aforementioned types, GMVAE learns to assign signals of the same type to the same cluster. Furthermore, as pure-noise signals tend to have lower energy compared to the corrupted signals, one can distinguish between the clusters by comparing the histograms of energies of the signals in each cluster. We note that this spectrum sensing approach is done in an unsupervised manner without the need for any labels.

\subsection{VAEs for PHY spoofing}\label{Sec:Applications to PHY spoofing}
We now study how VAEs can be leveraged to compromise the PHY security in NC-OFDM/OFDM systems.
As shown in Section \ref{sec:Learning useful representations from NC-OFDM signals}, VAEs can learn important information about the NC-OFDM transmission parameters in an unsupervised manner.
Specifically, a VAE extracts the following information about the NC-OFDM signals:
\begin{itemize}
\item \emph{Subcarrier occupancy pattern}: The learned representation $z_j$ corresponding to a subcarrier (correspondence can be found by latent traversal) that is inactive in a data sample $\mathbf{x}^i$ is close to zero as the power sent on that subcarrier is zero. Therefore, the subcarrier occupancy pattern can be inferred by applying a threshold test to the learned representations (threshold value depends on the SNR of the received signals and is estimated by cross-validation). 
\item \emph{Modulation scheme}: 
Through utilizing different modulation schemes in NC-OFDM signals, we have observed that the learned VAEs allocate one (two) latent variable(s) to each subcarrier when real (complex) symbols are being sent through each subcarrier. This can be justified by noting that VAEs treat the real and imaginary parts of a symbol transmitted through each subcarrier as distinct generative factors of the data and allocate separate latent variables to capture each one. This fact can be utilized to distinguish between modulation schemes that use real versus complex symbols (e.g., BPSK versus QAM). Also, as higher-order modulations send more power through each subcarrier, latent traversal can be utilized to distinguish them from lower-order ones. However, identifying the modulation scheme, in general, may need resorting to specific classifiers as discussed in \cite{FukingGunduz}.
\item \emph{Total number of active subcarriers}: As VAEs treat the real and imaginary powers in active subcarriers as generative factors of the dataset, the total number of active subcarriers in the whole dataset amounts to the number of informative latent variables when real symbols are being sent through each subcarrier (like in BPSK modulation), and is half this number when complex symbols (like in QAM or PSK modulation) are being used.
\end{itemize}
 For example, for the random allocation case considered in Fig. \ref{Fig:RandomAllocation}, we have observed that one latent variable encodes each subcarrier, which means BPSK is used as the modulation scheme, and there are $16$ informative latent variables, which indicates the number of subcarriers is $N=16$. It is shown in section \ref{Sec:Numerical examples} that VAEs are agnostic to the true signal bandwidth in these inferences. In other words, as long as signals are sampled above the Nyquist rate and stored in the dataset, the rate at which an adversary is sampling the received signals doesn't matter. This is particularly important from an adversary point of view because as pointed out in \cite{RatenshMilCom}, it is a major hurdle to estimate the bandwidth in the case of NC-OFDM signals.

We now incorporate these findings with the system model described in Section \ref{sec:SystemModel}. Here, the adversary only has access to a corrupted version of $s(t)$, denoted by $r(t)$ in Section \ref{sec:Channel Model}. As the training data samples are noisy in this case, we propose the following change to the original VAE objective function (\ref{eq:ELBO}):

\begin{equation}
\label{eq:ELBOnoisy}
\mathbf{\mathcal L}'(\mathbf x)=-\mathbb D_{KL}(q_{\phi}(\mathbf z|\mathbf x)||p(\mathbf z))+\eta\mathbb E_{q_{\phi}(\mathbf z|\mathbf x)}[\log p_{\theta}(\mathbf x|\mathbf z)],
\end{equation}
where we have weighted the reconstruction loss with a constant $\eta<1$ which is inversely proportional to SNR of the received signals and is obtained using cross-validation. This would lessen the reconstruction penalty and mitigate the effect of noise in the reconstructed signals by allowing VAEs to generate samples that are different than the input noisy ones. We note that the resulting bound remains a lower bound to the evidence $p(\mathbf x)$. We have seen through our experiments in Section \ref{Sec:Numerical examples} that this greatly improves the spoofing performance (which depends on the accuracy of the learned representations) in the case of noisy samples. After training a VAE on the dataset of received signals, during the test stage, the adversary inputs a signal to the trained VAE and estimates the aforementioned transmission parameters for PHY spoofing.


\section{PHY spoofing via supervised learning}\label{sec:Supervised Learning for PHY Spoofing}
\begin{figure}
        \centering
\includegraphics[scale=0.2]{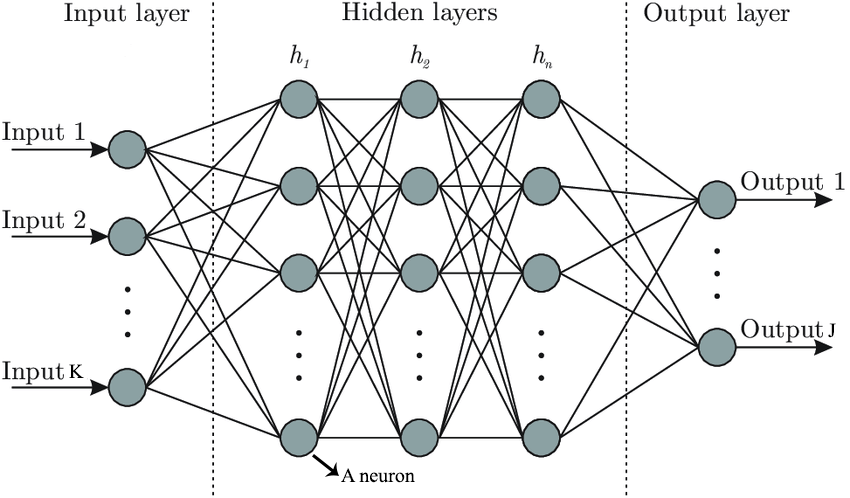}
\caption{Schematic of a DNN  with $3$ hidden layers.}
\label{fig:DNN}
\end{figure}

In this section, we assume the adversary has access to true labels for each data sample (i.e., a noisy received NC-OFDM/OFDM signal) in the dataset and makes use of them in the training stage. 
The adversary trains fully connected DNNs to estimate transmission parameters. Similar to the unsupervised spoofing, we assume that the adversary builds up a dataset out of the samples of the received noisy signals denoted by $r(kT_s)$ in Section \ref{sec:Channel Model} to extract $n_1$ (complex) samples. 
Fig. \ref{fig:DNN} demonstrates a general schematic of a DNN. Besides, Fig. \ref{fig:learning} illustrates two DNNs utilized for estimating transmission parameters, where the specifications of the upper DNN are:
\begin{figure}
    \centering

        \centering

     \includegraphics[scale=0.6]{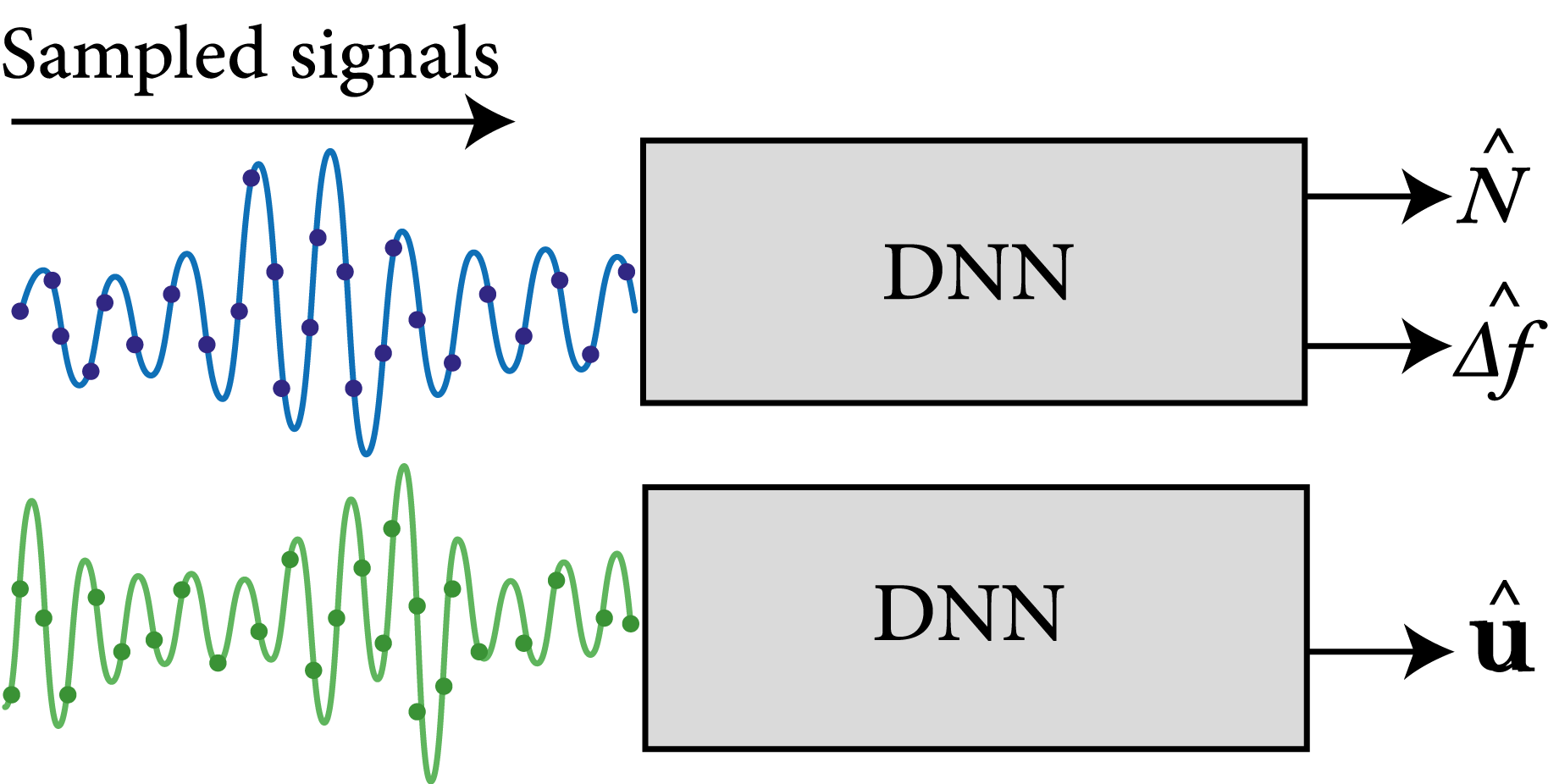}
\caption{Block diagrams of two DNNs used for estimating transmission parameters by the adversary.}
\label{fig:learning}
     \end{figure}

\begin{itemize}
\item {\bf Input: } $\mathbf{x}^i=[\mathcal R(r(0)),\dots, \mathcal R(r((n/2-1)T)), \mathcal I(r(0)),\dots, \mathcal I(r((n/2-1)T))]$.
\item {\bf Output: } Estimated total number of subcarriers and estimated subcarrier width, i.e., $[\widehat{N},\ \widehat{\Delta f}]$.
\item{\bf Architecture:}  See Table \ref{Tbl:DNNs}. The activation function for all the layers is chosen to be rectified linear unit (ReLU) function.
\item{\bf Training:} We minimize the $l_2$-loss $\lVert [\widehat{N},\ \widehat{\Delta f}]-[N,\ \Delta f] \rVert^2$, where $N$ and $\Delta f$ are the true parameters. Learning rate is set to $0.0001$.
\end{itemize}
The properties of the DNN utilized to infer $\mathbf{u}$ are described as follows:
\begin{itemize}
\item {\bf Input:} The same input vector described for the above DNN.
\item {\bf Output:} Estimated subcarrier occupancy pattern vector  $\widehat{\mathbf{\mathbf{u}}}$.
\item{\bf Architecture:}  See Table \ref{Tbl:DNNs}. The activation function for all the layers is chosen to be ReLU function except for the output layer, where the sigmoid function is used. The adversary further converts the output values to $0$'s and $1$'s using a hard-thresholding function defined by
$
h(t)=\begin{cases}
    0,& 0\leq t\leq 0.5,\\
    1,              & 0.5<t\leq 1.
\end{cases}
$
\item{\bf Training:} The DNN is trained by minimizing $Dist(\mathbf{\mathbf{u}},\widehat{\mathbf{\mathbf{u}}})=\sum_{i=1}^{N} (\mathbf{\mathbf{u}}(i)-\widehat{\mathbf{\mathbf{u}}}(i))^2$
where $\mathbf{\mathbf{u}}$ is the true subcarrier occupancy pattern. 
\end{itemize}
\begin{table}[]
\centering
\caption{\scriptsize Number of neurons in the hidden layers for the DNNs described in Fig. \ref{fig:learning}.}
\label{Tbl:DNNs}
\begin{tabular}{|l|l|l|l|l|}
\hline
  Hidden layer   & $1$ & $2$ & $3$ & $4$ \\ \hline
Upper DNN& $200$            & $400$            & $200$            & $50$             \\ \hline
Lower DNN& $350$            & $600$            & $400$            & $200 $           \\ \hline
\end{tabular}

\end{table}

\begin{figure}
    \centering

                   \includegraphics[width=7cm]{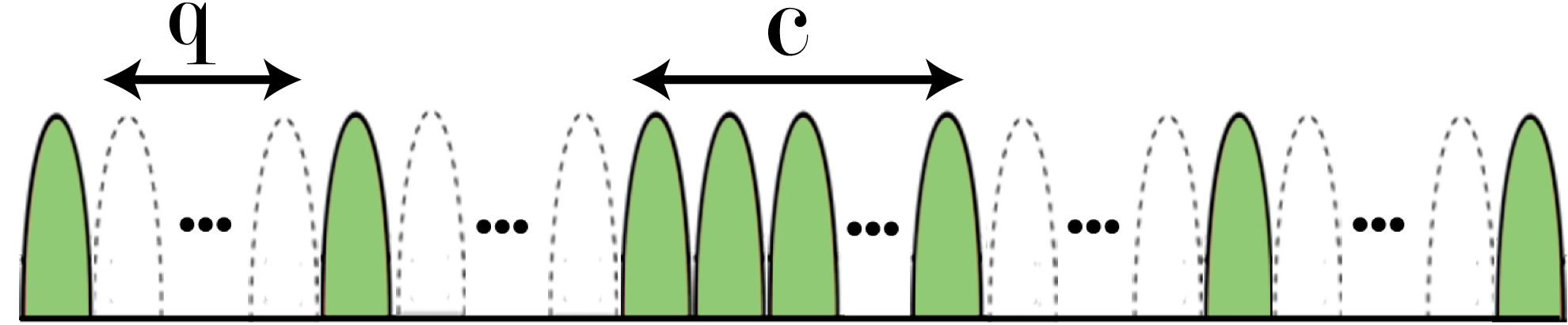}
\caption{Subcarrier occupancy pattern $1$; location of the contiguous block ($c$) is random.}
\label{fig:structNC1}
    \end{figure}

\begin{figure}
    \centering
        \includegraphics[width=8.6cm]{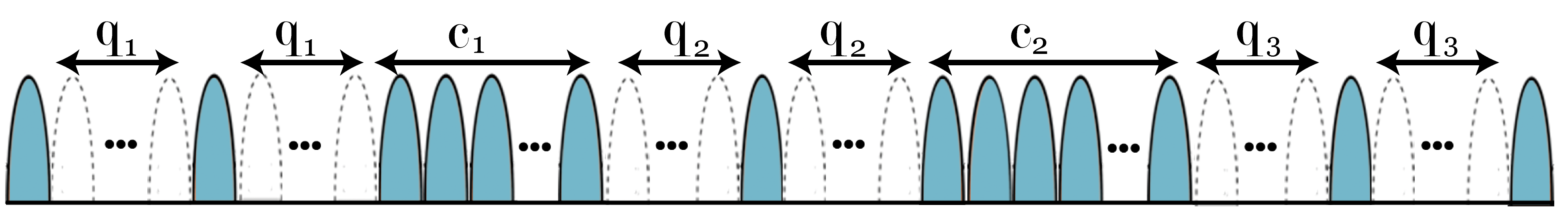}
\caption{Subcarrier occupancy pattern $2$; locations of the contiguous blocks are random.}
\label{fig:structNC2}
\end{figure}

We investigate the performance of the adversary when Tx is transmitting signals through $4$ different types of subcarrier occupancy patterns. $1$) A single contiguous block of active subcarriers (OFDM signal). $2$) NC-OFDM signal whose band allocation is illustrated in Fig. \ref{fig:structNC1} where integer $q$ denotes the number of inactive subcarriers between active ones and integer $c$ denotes the length of a block of contiguous active subcarriers. We refer to it as \textit{Pattern $1$} in the following. For training and test purposes, we generate signals of this type with $q$ in the range $[1,6]$ and $c$ in the range $[4,43]$ where the location of $c$ in the band is considered to be random. $3$) NC-OFDM signal whose band allocation is illustrated in Fig. \ref{fig:structNC2} and is referred to as \textit{Pattern $2$}. Here, there are two blocks of contiguous active subcarriers of length $c$, which is in the range $[3,15]$, and three different interleaved factors $q_1$, $q_2$ and $q_3$, all belonging to the range $[1,8]$. $4$) NC-OFDM signal where the bands are allocated in a random fashion without any specific pattern. In other words, we assume the transmitter flips a coin to decide whether a subcarrier is active or inactive.

The adversary receives signals of one of the above types at a certain SNR (spoofing SNR) during the training stage and trains the aforementioned DNNs on the corresponding dataset (one of the four datasets of signals). Then, it utilizes the trained model in the test stage to estimate the transmission parameters and PHY spoofs an unknown signal that has the same occupancy pattern type as in the training dataset. In this way, we are able to study how the choices of occupancy patterns affect the spoofing performance of the adversary. The number of signals in the training and test datasets for each case is set to $2\times10^6$ and $25\times10^4$, respectively. 
As mentioned in Section \ref{sec:SystemModel}, since the Tx may change the transmission parameters (particularly $\mathbf{\mathbf{u}}$) while transmitting an NC-OFDM symbol, the Rx also needs to infer these parameters, which can be done via DNNs as described above for the adversary. This is a fair model as the only advantage Rx has over the adversary is a better receiving channel. 


\subsection{Utilizing supervised learning to solve Example \ref{CAFexample}}\label{Sec:Utilizing DNN to solve the problem in Example}
We discussed in section \ref{sec:Signal exploitation using cyclostationary analysis} how CAF-based analysis fails to infer parameters of a simple NC-OFDM signal. Here, we now consider the same problem described in Example \ref{CAFexample} again, and utilize a DNN to solve it in a supervised manner. Here, as the underlying dataset is simpler (there are only $3$ different transmission cases) in comparison to the problem described in the previous section, we are able to solve the problem with a much simpler model. Specifically, we consider two architectures for the DNN. The first one has a hidden layer with $50$ neurons, and the second one has $3$ hidden layers of $500$, $250$ and $50$ neurons per each layer. The number of neurons at the input layer in both DNNs is set to $150$. The properties of the DNN which estimates the set of parameters $[q,T_u]$ are as follows.
\begin{itemize}
\item {\bf Input:} Samples of the received signal (similar to the previous DNNs).
\item {\bf Output:} Estimated parameters $[\hat{q},\widehat{T}_u]$.
\item{\bf Training:} We minimize the $l_2$-loss $\lVert [q,T_u]-[\hat{q},\widehat{T}_u] \rVert^2$, where $[q,T_u]$ are the true parameters.
\end{itemize}
As another approach, we consider a DNN which classifies between the three different transmission cases introduced in Table \ref{CAFTable}. The specifications for this DNN are as follows.
\begin{itemize}
\item {\bf Input:} Samples of the received signal (similar to the previous DNNs).
\item {\bf Output:} $\mathbf{\hat{p}}=[\hat{p}_1,\hat{p}_2,\hat{p}_3]$ where $\hat{p}_i$ denotes the estimated probability that the inputs correspond to case $i$, and $\sum_i \hat{p}_i=1$.
\item{\bf Training:} We minimize the cross-entropy loss between $\mathbf{\hat{p}}$ and $\mathbf{p}=[p_1,p_2,p_3]$, i.e., $-\sum_i p_i\log\hat{p}_i$ where $p_i$ represents the true probability that the signal belongs to case $i$.
\end{itemize}
For solving this problem, we consider training and test sets of size $5\times10^5$ and $3\times10^5$, respectively, which consist of the signals corresponding to the three transmission cases in Table \ref{CAFTable} at spoofing SNR of $5$ dB. The performances of these DNNs are presented in Section \ref{Sec:Numerical examples}, showing that they are indeed capable of estimating the transmission parameters that CAF had failed to infer.

\section{Disentanglement metrics and useful representations}\label{Sec:Disentanglement metrics}
Despite its importance, measuring the usefulness of the learned representation in a VAE model is not yet a well-studied subject. This is partly due to the context-dependent nature of the problem that makes it challenging for researchers to reach a consensus on this matter. As mentioned in Section \ref{sec:Unsupervised learning for PHY spoofing}, disentanglement is one of the features that is particularly desirable for the learned representations in a variety of applications. In this section, we characterize the properties of a disentangled representation by utilizing/extending the ideas from ML literature, describe why they are useful, and devise appropriate metrics to measure the disentanglement performance in the case of NC-OFDM datasets.
 Recall that \cite{FactorVAE} assumes a finite number of generative factors $k_1,\dots,k_f$ are used to generate the data samples in the dataset, where each factor could represent notions like position, scale, rotation, etc., in the case of image samples, or amount of power sent over distinct subcarriers for NC-OFDM/OFDM signals (see Section \ref{sec:Learning useful representations from NC-OFDM signals}). Then, \cite{FactorVAE} calls the learned representations disentangled if changes in one of the generative factors of the data are mirrored by the encoder $q_\phi(\mathbf z|\mathbf x)$ in Fig. \ref{Fig:VAE} in exactly one of the latent variables (i.e., one dimension of the latent space). We describe such an encoder to function in a disentangled manner. By extending this definition, we call the learned representations $\mathbf z$ disentangled if decoder $p_\phi(\mathbf x|\mathbf z)$ functions in a disentangled manner as well, i.e., it employs only one latent variable for each generative factor during the reconstruction process. In the remainder of this section, we discuss how to measure the disentangled performance of the encoder and decoder in VAEs utilizing existing and newly proposed metrics. 


In order to measure the performance of the encoder in achieving the disentangled representations, we will utilize the metrics proposed by Haggin et al. in \cite{BetaVAE} and by Kim et al. in \cite{FactorVAE}. The authors in \cite{BetaVAE} introduce a quantitative metric for measuring disentanglement assuming generative factors $k_1,\dots,k_f$ of the dataset are known. This metric works as follows. Choose a data sample $\mathbf{x}^0$ and obtains its corresponding representation $\mathbf{z}^0$ via the encoder of a trained VAE. Then, generate $L$ data samples $\{\mathbf{x}^i\}_{i=1}^{L}$ by fixing value of a generative factor $k_l$ ($l=1,\dots,f$) while changing values of the other factors uniformly at random in an interval (e.g. $[-3,3]$). Next, find their corresponding learned representations $\{\mathbf{z}^i\}_{i=1}^{L}$, and use $\frac{1}{L}\sum_{i=1}^{L}|\mathbf{z}^i-\mathbf{z}^0|$ and the fixed factor's index ($l$) as a training sample to train a linear classifier where $l$ is treated as label. Then, the correct classification rate achieved by this classifier on a test set (generated in the same way as the training set) is reported as the metric. 
A more stringent metric for disentanglement is proposed in \cite{FactorVAE}, which works as follows. Choose a generative factor ($k_l$), generate data $\{\mathbf{x}^i\}_{i=1}^{L}$ with this factor fixed while the others' values are changed randomly, and obtain their corresponding learned representations $\{\mathbf{z}^i\}_{i=1}^{L}$. Then, normalize each dimension of  $\mathbf{z}^i,\ i=1,\dots,L$, by its empirical standard deviation computed over $\{\mathbf{z}^i\}_{i=1}^{L}$ to obtain normalized representations $\{\mathbf{\bar{z}}^i\}_{i=1}^{L}$. The empirical variance for an arbitrary real-valued vector $\mathbf w=[w_1,w_2,\dots,w_h]$ of length $h$ is defined as
\begin{equation}
    \widehat{Var}(\mathbf w)=\frac{1}{2h(h-1)}\sum_{i,j=1}^{h}(w_i-w_j)^2.
\end{equation}
Then, compute the empirical variances of each dimension in $\{\mathbf{\bar{z}}^i\}_{i=1}^{L}$, which are denoted by $[v_1,\dots,v_{N_{\mathbf{z}}}]$. Afterward, the index of the dimension with the lowest variance ($\argmin_i\ v_i$) and the fixed factor index $l$ provide one training sample for the classifier. The performance of this classifier (measured between $0$ and $100$) on the test set is reported as the final metric.

Next, we evaluate the performance of the VAE's decoder in achieving disentangled representations utilizing latent traversal method described in Section \ref{sec:Learning useful representations from NC-OFDM signals}. 
In the case of disentangled representations, each latent variable governs a specific generative factor in the reconstructed sample through the decoder.
Algorithm \ref{Alg:ourmetric2} shows the pseudo-code for our proposed metric, which outputs a decimal value between $0$ and $100$. Specifically, this algorithm changes the value of the $j$-th latent variable $z^i_j$ (corresponding to the input data sample $\mathbf x^i$) in range $[-C,C]$ with steps of $2C/K$ while fixing the others, obtains the element-wise difference between FFT of the reconstructed samples and FFT of $\mathbf x^i$, and counts the number of subcarriers in which the magnitude of the difference is more than a predefined threshold $\epsilon$. If there is only one subcarrier, it represents a disentangled latent variable and $z^i_j$ gets a perfect score ($100$). If there is more than one, then $z^i_j$ gets $0$. This would be done for all the $N_{\mathbf{z}}$ latent variables, and $L$ different data samples \{$\mathbf x^i\}_{i=1}^L$. Then, the final metric equals the averaged score over all the data samples and the number of informative latent variables (denoted by $I$ in Alg. \ref{Alg:ourmetric2}). We have used $C=3, K=40, L=500$ (larger values did not change the result noticeably) and two different values of $\epsilon$, i.e., $0.5$ and $1$ in our experiments.

We consider a VAE to have learned disentangled representations if it achieves a score of $100$ for both Kim's metric \cite{FactorVAE} and the proposed latent traversal metric (Alg. \ref{Alg:ourmetric2}). Such a VAE is well-suited for a dataset of NC-OFDM signals where the amount of power is independent in different subcarriers. More specifically, as VAEs encode these power in the latent space $\mathbf{z}$, the corresponding learned representation is expected to be disentangled. Therefore, designing a VAE model that captures this important property of the dataset enables one to find accurate representations and improve the performance of the unsupervised PHY spoofing described in Section \ref{sec:Unsupervised learning for PHY spoofing}. 

\begin{algorithm}
\small
\SetAlgoLined

\textbf{Output:} Disentanglement metric $S0$ (out of $100$) and $I$, the number of informative latent variables.\\
\textbf{Input:} A VAE model ($\phi$, $\theta$), dataset $\{\mathbf{x}^i\}_{i=1}^{L}$, number of latent variables $N_{\mathbf z}$, traversing limit $C$, traversing step $K$, averaging factor $L$, precision factor $\epsilon$, $S1=0$ .\\
\For{$i=1$ \textbf{to} $L$}{
Pick data point ($\mathbf{x}^i$),
$S2\gets0$, $I\gets0$\\
 \While{$j<N_{\mathbf z}+1$}{
 $\mathbf{z}^i\gets$ mean of $q_\phi(\mathbf z|\mathbf x^i), \mathbf{v}_1\gets$ zero-vector, $j\gets1$\\
  \For{$k=0$ \textbf{to} $K$}{
  $\mathbf{z}^i_j\gets -C + 2Ck/K$, $\mathbf{v}_2\gets$ zero-vector,\\  $\widehat{\mathbf{x}}\gets$ mean of $p_\theta(\mathbf x|\mathbf z^i)$\\
  $\mathbf{F}\gets$ Element-wise magnitude of $(FFT(\hat{\mathbf{x}})-FFT(\mathbf{x}^i))$, $T\gets$ length of $\mathbf{F}$\\$\mathbf{v}_2(t)\gets1$ if $\mathbf{F}(t)>\epsilon$ for $t=1,\dots,T$,\   $\mathbf{v}_1\gets \mathbf{v}_1+\mathbf{v}_2$\\

   }
   $k\gets$ Number of non-zero elements in $\mathbf{v}_1$\\
  \uIf{$k==1$}{
  $S2\gets S2+100,$ $I\gets I+1$
   }\uElseIf{$k>1$}{
        $S2\gets 0$, $I\gets I+1$\\
  }
  $j\gets j+1$
}
$S1\gets S1+S2/I$
 }
 $S0=S1/M$
 \caption{Latent traversal for a dataset of NC-OFDM signals}
 \label{Alg:ourmetric2}
\end{algorithm}

\section{Numerical examples}\label{Sec:Numerical examples}
\begin{figure}

\centering
        \includegraphics[width=7.5cm]{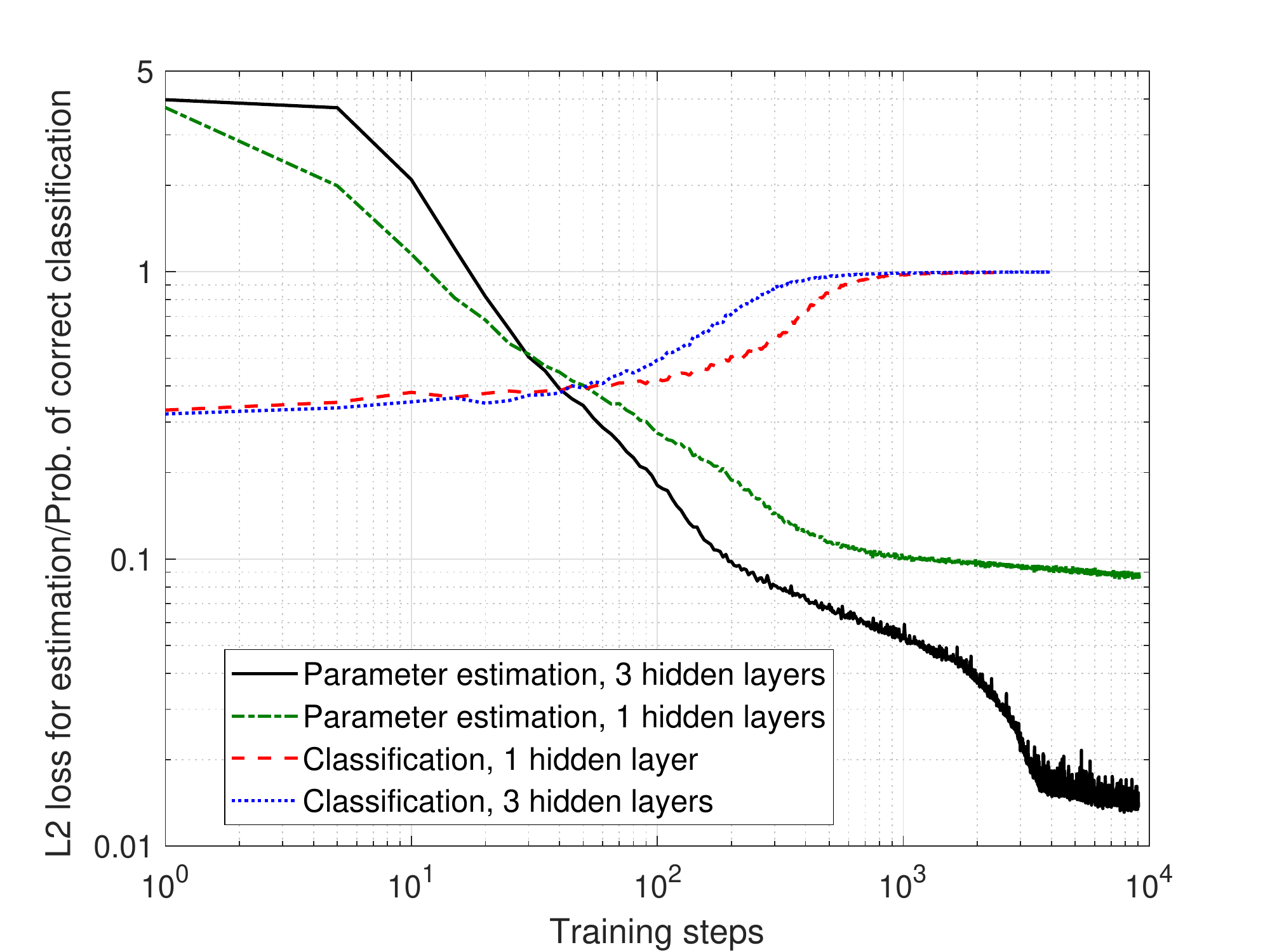}
        \caption{Performance of DNNs described in Section \ref{Sec:Utilizing DNN to solve the problem in Example} in solving Example \ref{CAFexample}.
        \label{Fig:CAFCOMP}}

\end{figure}

In this section, we first present the result of numerical simulations that characterize the performance of an adversary utilizing unsupervised and supervised learning algorithms for PHY spoofing as discussed in Sections \ref{sec:Unsupervised learning for PHY spoofing} and \ref{sec:Unsupervised learning for PHY spoofing}. Then, we provide our results and insights on disentanglement metrics presented in Section \ref{Sec:Disentanglement metrics}. We begin with investigating the performance of the supervised learning based on DNNs described in Section \ref{Sec:Utilizing DNN to solve the problem in Example} for spoofing the system introduced in Example \ref{CAFexample}. This is demonstrated in Fig. \ref{Fig:CAFCOMP} where two different DNN structures are considered for each classification and estimation scenario. Note that the $y$-axis represents correct classification probability for classification curves, and $l_2$-loss for the estimation problems on the test set. Specifically, it is shown that for the classification problem, even using one hidden layer enables us to identify the true class of the test signals after around $1000$ training steps with very high probability. In each training step, mini-batches of size $500$ are chosen from the training set and the SGD algorithm with a learning rate of $0.0005$ is applied to train the DNNs. Furthermore, if the adversary wishes to estimate the true parameters of the signal (i.e., $[q,T_u]$), average $l_2$-losses as low as $0.1$ are achievable using DNN estimator with only one hidden layer. We also observe that a higher number of layers results in better performances in both cases.
 \begin{figure}
        \centering
        \includegraphics[width=7.5cm]{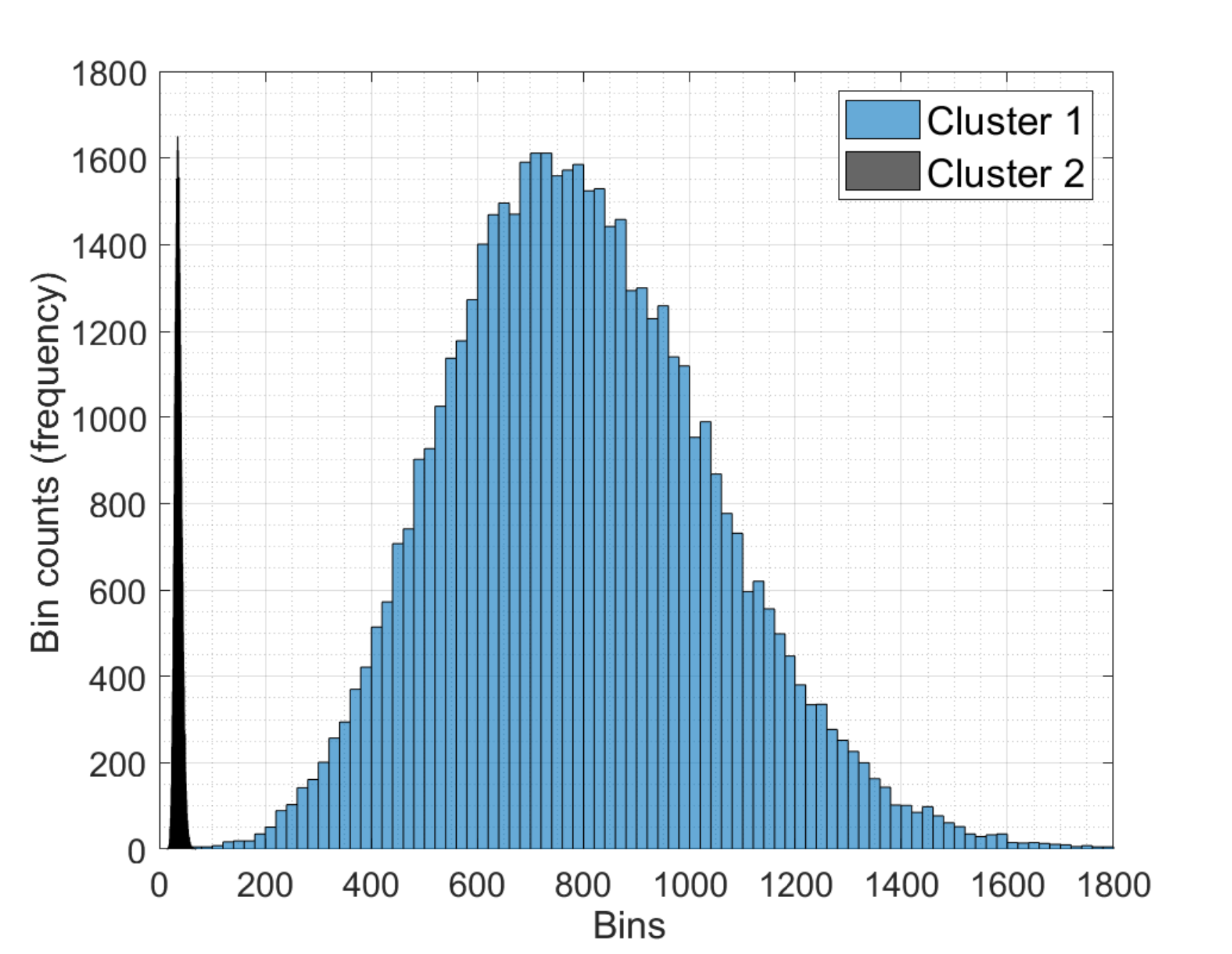}
        \caption{Histogram of energies of signals in two clusters of GMVAE.}
        \label{fig:Hist}

\end{figure}

Now, we consider the system model described in Section \ref{sec:SystemModel} where the corresponding PHY parameters between different communication parties follow the narrowband
(NB)-IoT \cite{NBIOT1,NBIOT2} standard that is widely used in the existing IoT solutions \cite{NBIOT1}. Specifically, the transmission is assumed to take place over an NC-OFDM scheme with $\Delta f\in[15,30,45,60]$ KHz and $p_n\in[1,2]$ utilizing BPSK modulation with random subcarrier occupancy pattern using the total number of subcarriers $N=16$ and $N=32$, which give rise to $2^{16}$ and $2^{32}$ distinct subcarrier occupancy patterns, respectively (see Section \ref{sec:Learning useful representations from NC-OFDM signals}). Adversary overhears the transmissions at a certain \textit{spoofing SNR} and builds up a dataset out of the received noisy signals where $80$ complex samples are collected from each signal. The size of the training and the test dataset is set to $2\times10^6$ and $25\times10^4$, respectively. As described in Section \ref{sec:Spectrum sensing via VAEs}, the adversary first trains a GMVAE on a dataset consisting of both pure noise signals and corrupted NC-OFDM signals in order to distinguish between the two. Specifically, we have trained a GMVAE with $k=2$, where the dimension of both $\mathbf{z}$ and $\mathbf{w}$ is set to $20$, on a dataset consists of $50000$ pure noise and noisy NC-OFDM signals at SNR $5$ dB with random band allocation and $N=16$. Fig. \ref{fig:Hist} shows the histograms of energies of the signals in each cluster, where a clear distinction between the two is demonstrated. As noise signals have lower energy, one would know that they will belong to cluster $2$. The classification accuracy of GMVAE is evaluated as $99.6\%$.

\begin{figure}
        \centering
         \begin{minipage}{0.45\textwidth}
        \includegraphics[width=1\textwidth]{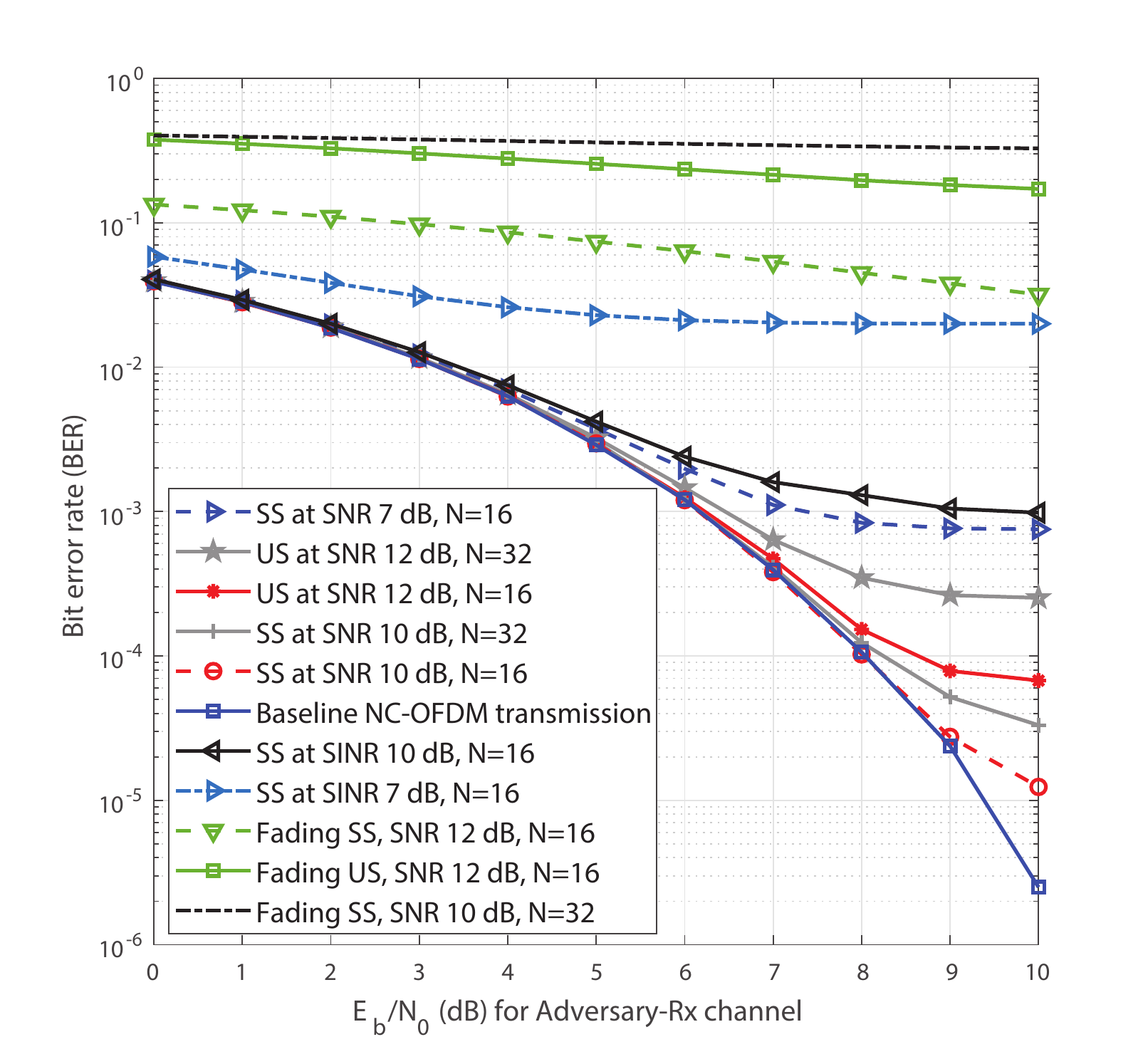}
        \caption{Spoofing performance of the adversary in different scenarios. SS and US denote supervised and unsupervised spoofing, respectively.}
        \label{Fig:N16spoofing}
    \end{minipage}\hfill
    \begin{minipage}{0.45\textwidth}
        \centering
        \includegraphics[width=1\textwidth]{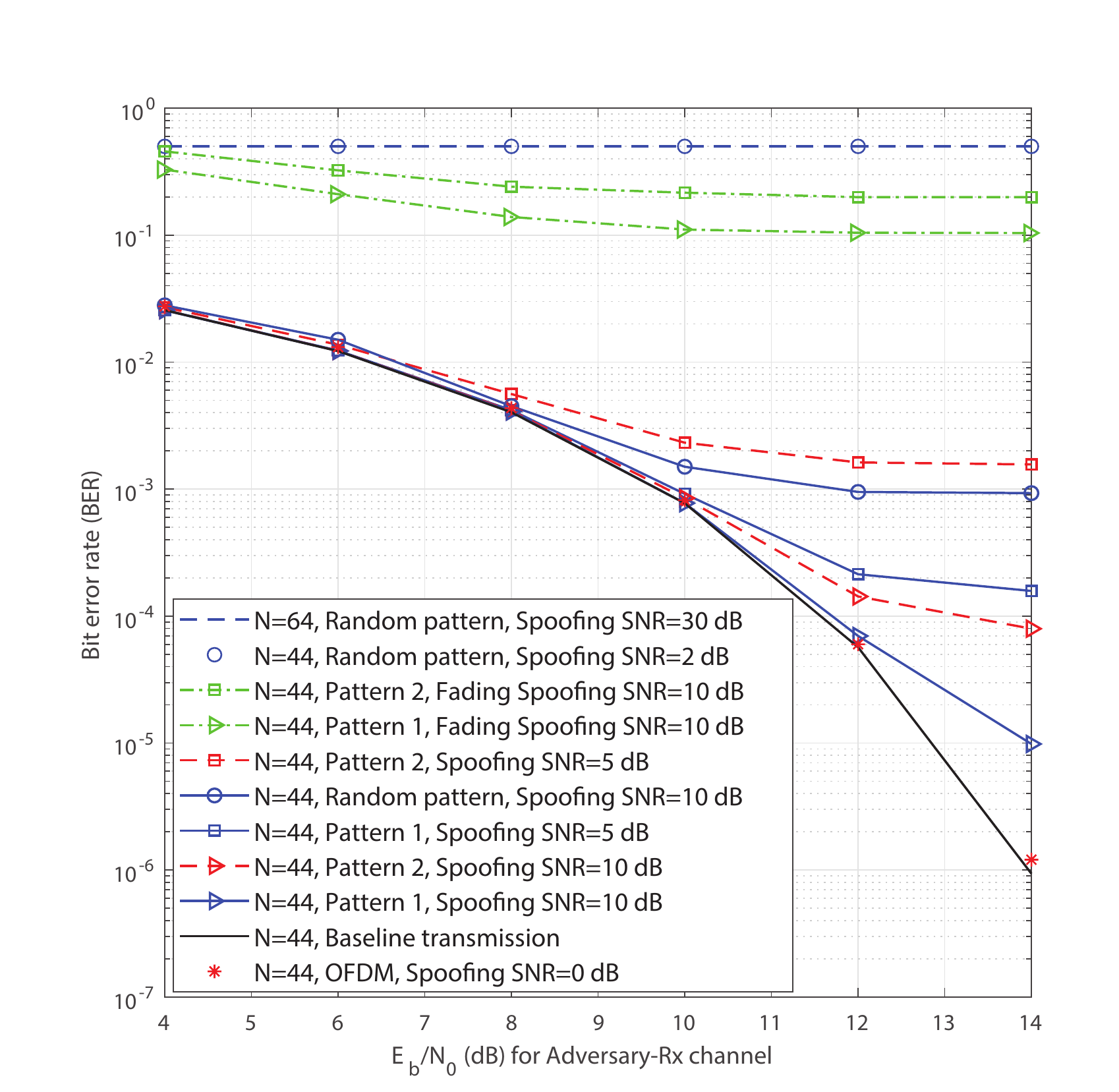}
        \caption{Supervised spoofing performance of the adversary when different frequency occupancy patters are used by the Tx.}
        \label{fig:N64Spoofing}
    \end{minipage}

\end{figure}

Fig. \ref{Fig:N16spoofing} illustrates the spoofing performance for several supervised and unsupervised learning algorithms for both AWGN and fading environments. For the fading case, we consider a multi-path fading channel with amplitudes $[1,0.8,0.6]$ and delays $[0,2,4]\mu s$ for Tx-adversary channel, and a Rayleigh flat-fading for both of the Adversary-Rx and Tx-Rx channels. For the unsupervised cases, we assume the adversary is utilizing the FactorVAE model with $\gamma=5$. Then, it infers the total number of subcarriers and the corresponding latent variable for each subcarrier via latent traversal as described in Section \ref{sec:Unsupervised learning for PHY spoofing}. During the test stage, it obtains the corresponding learned representation (mean of $q_\phi(\mathbf z|\mathbf x)$) for a test signal and decides whether a subcarrier is active or inactive as discussed in Section \ref{sec:Unsupervised learning for PHY spoofing}. For the supervised cases, the specifications of the DNNs are presented in Section \ref{sec:Supervised Learning for PHY Spoofing}. 
Fig. \ref{Fig:N16spoofing} shows that the supervised algorithm offers better spoofing performance in general as it relies on the ground truth labels. Spoofing performance in a fading channel is worse for both supervised and unsupervised algorithms in comparison to the AWGN channel as the quality of the received signals by the adversary is deteriorated. This also holds for the case where the signals in the dataset are corrupted by the interference from other transmitting sources at different signal to interference plus noise ratios (SINRs). The higher number of subcarriers $N$ ($32$ vs. $16$), in general, negatively affects the performance of the adversary in both cases. Also, as the spoofing SNR increases, spoofing performance improves as well. Specifically, if the spoofing SNR is $10$ dB, the supervised spoofing performance can get very close to the baseline transmission (the best spoofing performance). Furthermore, it is shown that the adversary can spoof the signal to certain degrees via VAEs if the spoofing SNR is high (e.g., $12$ dB). We note that as the adversary estimates the transmission parameters with higher accuracy, it can achieve better spoofing performance. For example, for the supervised spoofing in the AWGN scenario with $N=16$ presented in Fig. \ref{Fig:N16spoofing}, mean error estimation of subcarrier occupancy pattern by the adversary decreases from $1.2\times 10^{-3}$ to $1.6\times 10^{-4}$ as spoofing SNR increases from $7$ dB to $10$ dB. Similarly, for unsupervised spoofing at SNR $10$, this error is computed as $8\times 10^{-4}$ and $4.5\times 10^{-4}$ for $N=32$ and $N=16$, respectively, which indicates a better spoofing performance in the latter case.  

Next, we consider an NC-OFDM transmission with a higher number of subcarriers $N=44$ where $16$-QAM modulation is utilized by the Tx. We investigate the performance of the adversary utilizing the supervised learning algorithm for spoofing in both AWGN and fading environments, and assume it collects $n_1=100$ samples from each received signal at different spoofing SNRs. Here, we consider $4$ different occupancy patterns described in Section \ref{sec:Supervised Learning for PHY Spoofing} to investigate their resilience against PHY spoofing. Fig. \ref{fig:N64Spoofing} demonstrates the performance of the adversary in different scenarios based on different spoofing SNRs and occupancy pattern vectors. For the OFDM scenario, the adversary is able to infer the occupancy pattern without error and can achieve the same performance as the baseline transmission. For Pattern $1$ NC-OFDM signal, the DNNs are trained and tested at $3$ different spoofing SNRs of $5$ and $10$ dB, where as expected the higher spoofing SNRs result in better spoofing performance. Comparing to the baseline transmission, there is a gap in the performances since the parameters are estimated with some error.
For Pattern $2$, one can see that the performance is worse in comparison to Pattern $1$ as the Tx is using a more complex subcarrier occupancy pattern with a higher number of distinct band allocations.
Therefore, one can conclude that Pattern $2$ NC-OFDM signals are more difficult to spoof. It is also shown that the spoofing performance is worse in fading scenarios (with the same specifications described for Fig. \ref{Fig:N16spoofing}) where the signals are received from a multipath channel.

For the random occupancy case, it is shown that BER is $0.5$ at spoofing SNR of $2$ dB for a wide range of $E_b/N_0$'s corresponding to the adversary-Rx channel, which indicates that the adversary is unable to spoof such signals at SNR $2$ dB (or lower) because of the high estimation errors at the output of the DNNs. As a result, the PHY is considered to be secure when spoofing SNR is lower than $2$ dB while Tx-Rx SNR is set to $16$ dB in order to ensure reliable communication in the meantime. Fig. \ref{fig:N64Spoofing} also demonstrates the spoofing performance for the case of random occupancy patterns with $N=64$. Due to the larger number of possible subcarrier occupancy patterns, the Rx is unable to estimate the transmission parameters via the Tx-Rx channel correctly even when the Tx-Rx SNR is $30$ dB for this case which precludes achieving a reliable communication. In other words, although this scheme prevents the adversary from PHY spoofing, it also fails to ensure reliable communication for legitimate Tx and Rx pairs. In the case where multiple users are transmitting over a certain frequency band in the presence of spectrum incumbents, random band allocation, which is now applied to the accessible frequencies only, enables opportunistic use of the spectrum holes. This intrinsically improves spectral efficiency and enables shared multiple access for the users. Even if there is no spectrum incumbent, the NC-OFDM scheme while providing secure transmission can recover the loss of spectrum efficiency when multiple links, each using NC-OFDM, share the available bandwidth as shown in \cite{minsum}.

\begin{figure}

        \centering

        \includegraphics[width=6.5cm]{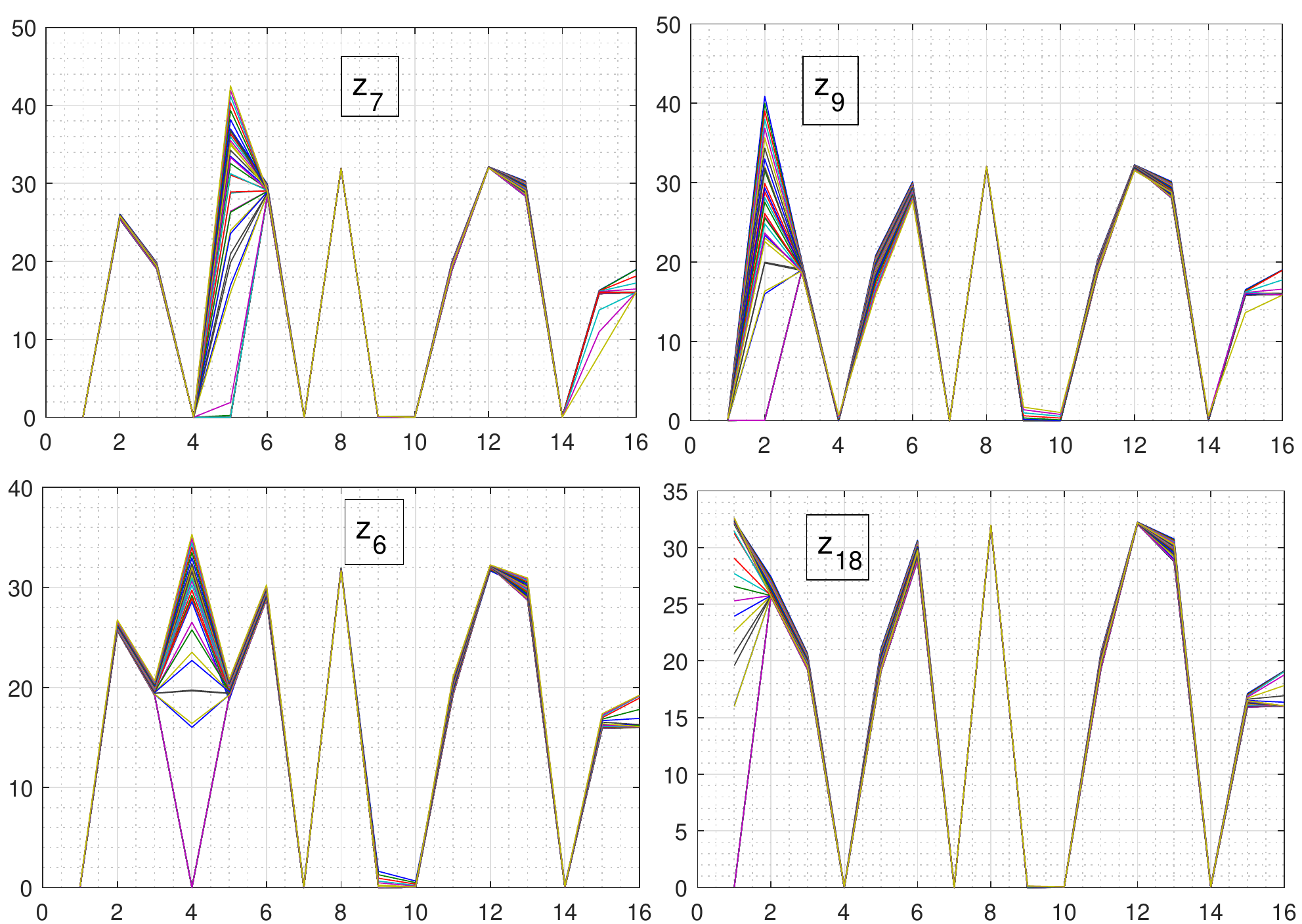}

        \caption{FFT of the reconstructed signals in DIP-VAE ($\lambda_d=\lambda_{od}=50$) obtained under latent traversal. $z_i$ denotes the fixed latent variable. 
        }
        \label{Fig:DIP50}

    \end{figure}
\begin{figure}

        \centering

        \includegraphics[width=6.5cm]{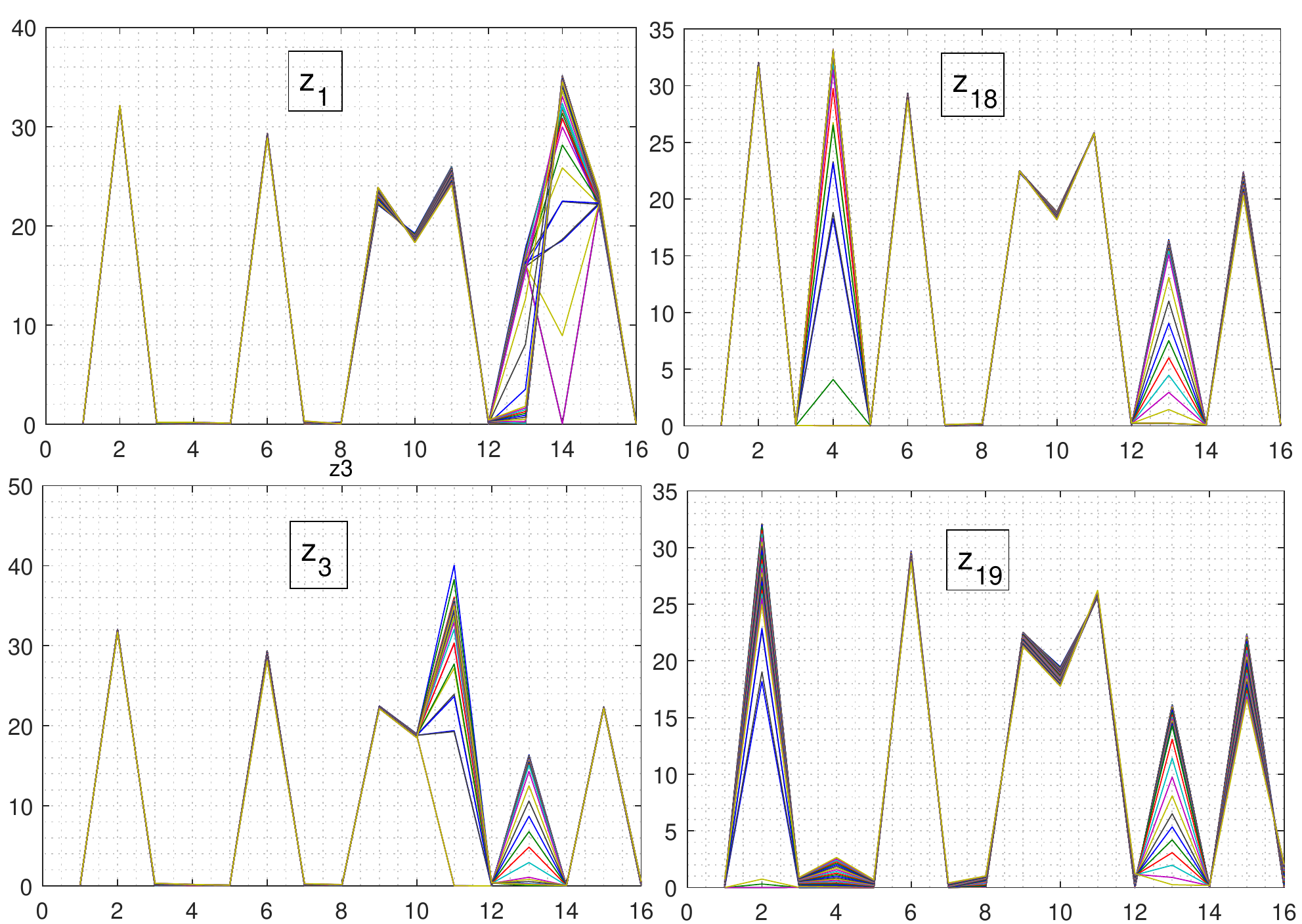} 

        \caption{FFT of the reconstructed signals in DIP-VAE model ($\lambda_d=\lambda_{od}=5$) obtained under latent traversal. 
        }
        \label{Fig:DIP5}

\end{figure}

Next, we have evaluated the metrics described in Section \ref{Sec:Disentanglement metrics} in Table \ref{Tbl:VAEMetrics} for different VAE models trained on a dataset of NC-OFDM signals with $N=16$ and random subcarrier occupancy patterns where BPSK modulation with $p_n\in[1,2]$ is used. The number of latent variables is set to $N_{\mathbf{z}}=20$ and the encoder/decoder is modeled with feed-forward DNNs described in Section \ref{sec:Learning useful representations from NC-OFDM signals}. Table \ref{Tbl:VAEMetrics} shows that the performance of the encoder and decoder in terms of disentanglement may vary greatly. In other words, metrics in \cite{BetaVAE,FactorVAE} do not guarantee a disentangled representation in the case of latent traversal. For example, for the DIP-VAE models described in Section \ref{sec:Unsupervised learning for PHY spoofing} (with parameters $\lambda_d=\lambda_{od}=5$ and $\lambda_d=\lambda_{od}=50$), one can see that although they achieve perfect scores ($100$) for the metrics proposed in \cite{BetaVAE,FactorVAE}, they perform poorly on the metric based on the latent traversal. This is illustrated in Figs. \ref{Fig:DIP50} and \ref{Fig:DIP5}.
It is clear that the learned representation could not be considered disentangled since changing one latent variable affects the values of the others. Reconstruction error is also reported in Table \ref{Tbl:VAEMetrics}, which represents the ability of the VAE model to reconstruct the input signal $\mathbf x^i$, and is defined as $\sum_{j=1}^d|\mathbf{x}^i(j)-\mathbf{\hat{x}^i}(j)|$ (element-wise subtraction) where $d$ denotes the dimension of $\mathbf x^i$.

\begin{figure}
\centering

        \centering
        \includegraphics[width=7cm]{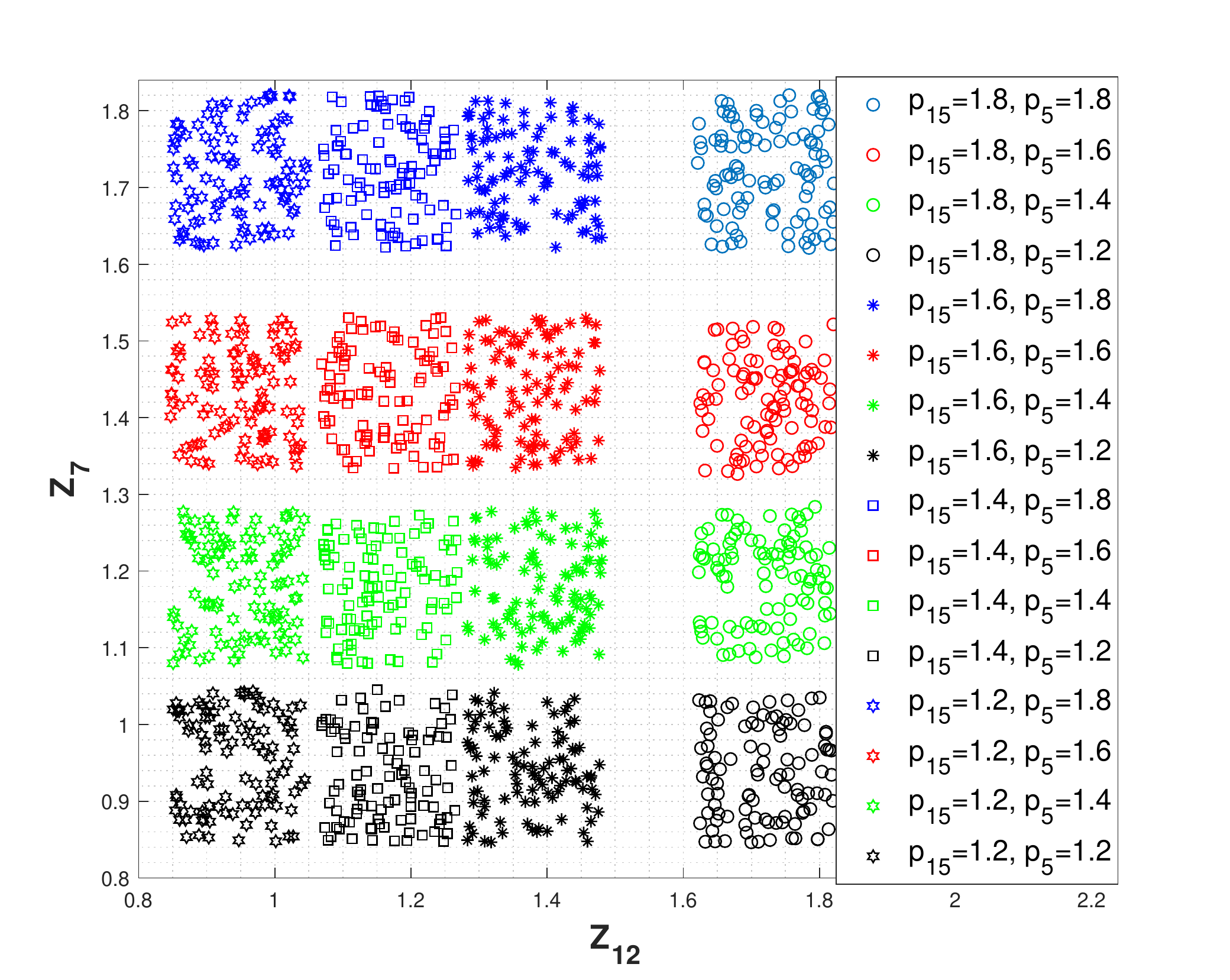}
        \caption{The learned representations $z_{12}$ and $z_7$ for FactorVAE model ($\gamma=5$) which control the amount of power in subcarriers $15$ and $5$ ($p_{15}$ and $p_{5}$) of the reconstructed signal, respectively, .}
        \label{Fig:Intuition-Good}
  \end{figure}

\begin{figure}
        \centering
         \includegraphics[width=7cm]{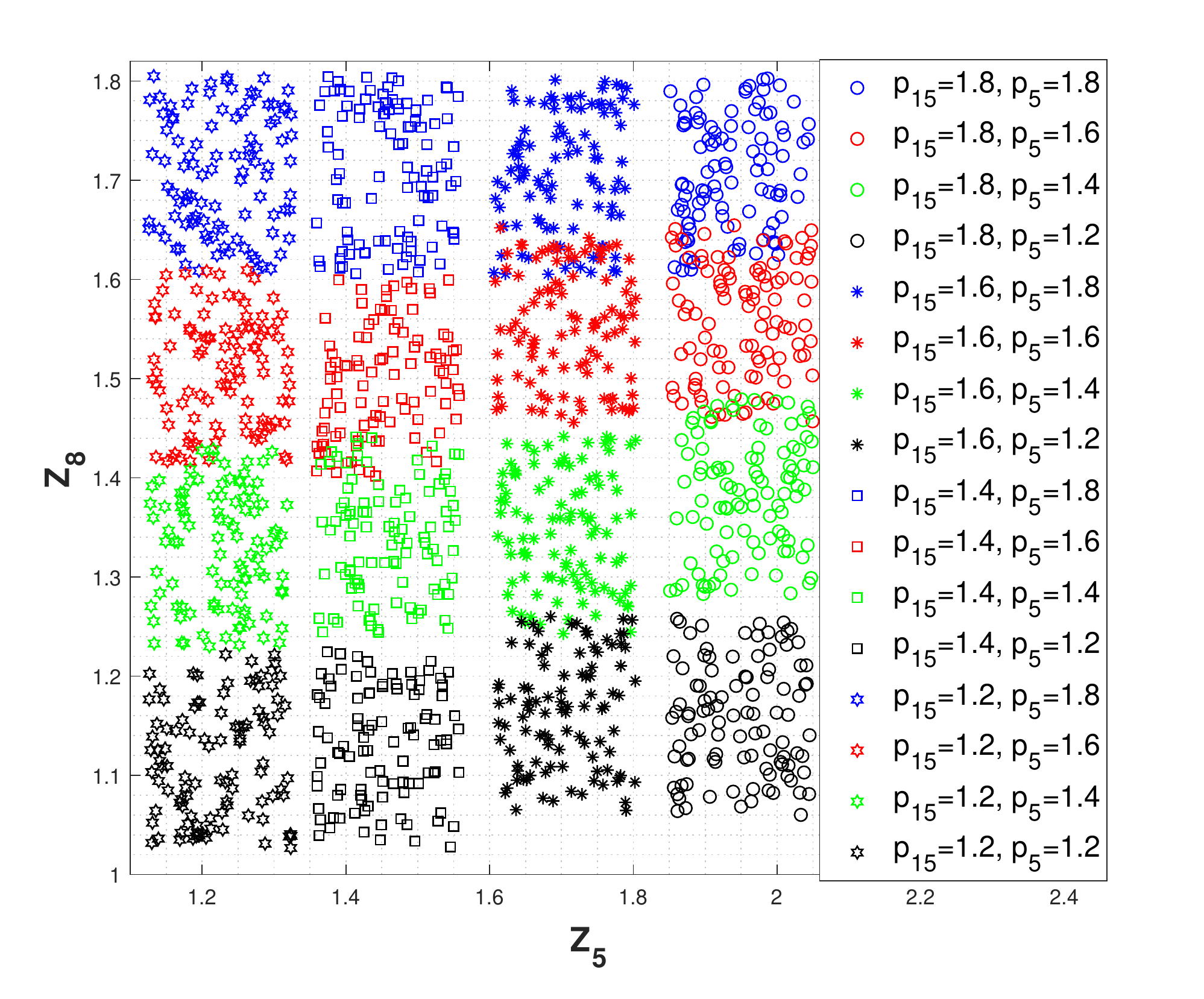} 

        \caption{The learned representations $z_{5}$ and $z_{8}$ for DIP-VAE model ($\lambda_d=\lambda_{od}=50$) which control the amount of power in subcarriers $15$ and $5$ ($p_{15}$ and $p_{5}$) of the reconstructed signal, respectively, }
        \label{Fig:Intuition-Bad}

 \end{figure}

\begin{table*}
\centering
\caption{\scriptsize Disentanglement metrics and reconstruction loss computed for different VAE models }

\label{Tbl:VAEMetrics}
\begin{tabular}{|l|l|l|l|l|l|l|}
\hline
Method & Parameters  & \begin{tabular}[c]{@{}l@{}}Haggin's\\ metric \cite{BetaVAE}\end{tabular} & \begin{tabular}[c]{@{}l@{}}Kim's\\ metric \cite{FactorVAE}\end{tabular} & \begin{tabular}[c]{@{}l@{}}Alg. \ref{Alg:ourmetric2} metric \\ ($\epsilon=0.5$)\end{tabular} &\begin{tabular}[c]{@{}l@{}}Alg. \ref{Alg:ourmetric2} metric \\ ($\epsilon=1$)\end{tabular} & \begin{tabular}[c]{@{}l@{}}Reconst.\\ error\end{tabular} \\ \hline
$\beta$-VAE \cite{BetaVAE}& $\beta=10$ & 100 & 100 & 98.25 &100 &6.5e-2\\ \hline
$\beta$-VAE & $\beta=20$ & 99 & 92 & 99.75&100 &8.3e-2\\ \hline
DIP-VAE \cite{DIPVAE}& $\lambda_d=\lambda_{od}=5$ & $100$ & $100$ & $59.37$ & $84.06$&$1.9e-2$\\ \hline
DIP-VAE & $\lambda_d=\lambda_{od}=10$ &  $100$ & $66$ & $85.5$ & $98.43$&$2.3e-2$\\ \hline
DIP-VAE & $\lambda_d=\lambda_{od}=50$ &  $100$ & $100$ & $10.2$ & $25.06$& $2.3e-2$\\ \hline
Factor-VAE \cite{FactorVAE}& $\gamma=5$ &  $100$ & $100$ & $97.62$ & $99.93$&$1.9e-2$\\ \hline
Factor-VAE & $\gamma=10$ & $95$ & $40$ & $70$ & $85.75$ &$3e-2$\\ \hline
Factor-VAE & $\gamma=50$ & $98$  & $66$ & $20.75$ &  $72.18$&$3.3e-2$\\ \hline
\end{tabular}
\end{table*}
As discussed in Section \ref{Sec:Disentanglement metrics}, we say a VAE model results in perfectly disentangled representations if it achieves perfect scores for both Kim's metric \cite{FactorVAE} and the latent traversal metric (Alg. \ref{Alg:ourmetric2}). There, we also argued how such a VAE is particularly relevant for PHY spoofing NC-OFDM/OFDM signals where the amount of power is independent in different subcarriers. Specifically, the Kim's metric ensures that a specific learned representation (by the encoder) corresponding to a generative factor is not affected by the changes in other generative factors. The latent traversal metric implies that the learned representations corresponding to different generative factors are disjoint. This can be seen by comparing the learned representations for two different VAEs: DIP-VAE with ($\lambda_d=\lambda_{od}=50$) and FactorVAE with $\gamma=5$ whose latent traversal performance is demonstrated in Figs. \ref{Fig:DIP50} and \ref{Fig:RandomAllocation}, respectively. First, it should be noted that only in FactorVAE model, which achieves perfectly disentangled representations, a latent variable corresponds to exactly one subcarrier via the decoder. For these VAEs, we also have illustrated the learned representations corresponding to two different latent variables via a $2$D space in Figs. \ref{Fig:Intuition-Good}. The VAE's decoder maps each point in this space to a signal whose corresponding power in two different subcarriers follows the values of these two latent variables. Both models achieve perfect scores on Kim's metric \cite{FactorVAE}. However, one can see that the learned representations for the case of FactorVAE in Fig. \ref{Fig:Intuition-Good} are disjoint while they overlap for DIP-VAE in Fig. \ref{Fig:Intuition-Bad}. In fact, this is the reason why DIP-VAE in Fig. \ref{Fig:DIP50} gets a low score on the latent traversal metric. By changing the value of one of the latent variables (e.g. $z_5$) at a time, a signal can be generated whose power changes in more than one subcarrier ($p_5$ and $p_{15}$). On the other hand, in a perfectly disentangled latent space, the learned representations that get mapped to signals with different power levels in each subcarrier form a disjoint region as illustrated in Fig. \ref{Fig:Intuition-Good}. Specifically, one can see that the representations corresponding to a signal with specific power levels in different subcarriers form a region in $\mathbf z$ space. Disjoint representations facilitate downstream tasks like classification as the data samples generated by different generative factors get mapped to distinct regions.




\section{Conclusions}\label{Sec:Conclusions}
We have investigated the PHY robustness of the NC-OFDM/OFDM system in IoT against an adversary equipped with machine learning tools. Specifically, we have assumed the adversary employs supervised and unsupervised learning algorithms to infer some of the NC-OFDM transmission parameters and physically spoof the system. The proposed unsupervised algorithm utilizes VAEs for spoofing, which can infer important spectral information about the NC-OFDM/OFDM signals.
Numerical results demonstrate that the PHY spoofing performance highly depends on the subcarrier occupancy pattern used by the transmitter. Specifically, the results suggest that the transmitter should randomize the selection of the active subcarriers in order to impede PHY spoofing attacks utilizing DNNs.


\begin{IEEEbiography}[{\includegraphics[width=1in,height=1.25in,clip,keepaspectratio]{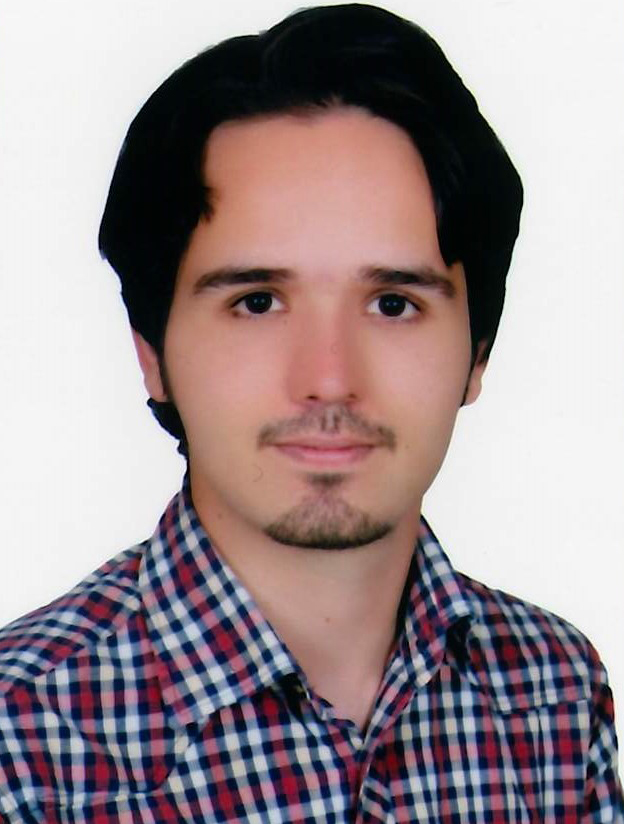}}]{Alireza Nooraiepour} earned the B.S. degree in Electrical Engineering from Amirkabir University of Technology (Tehran Polytechnic) in 2013, and the M.S. degree in Electrical and Electronics Engineering from Bilkent University in 2016. He is now pursuing his Ph.D. degree in WINLAB, Rutgers University, NJ, USA. His current research focuses on deep learning, internet of things and physical layer security.
\end{IEEEbiography}
 \vskip 0pt plus -1fil
\begin{IEEEbiography}[{\includegraphics[width=1in,height=1.25in,clip,keepaspectratio]{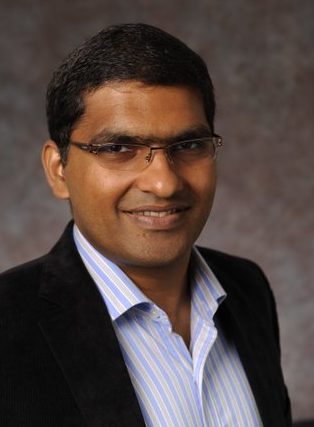}}]{Waheed U. Bajwa}  (Senior Member, IEEE) received BE (with Honors) degree in electrical engineering from the National University of Sciences and Technology, Pakistan in 2001, and MS and PhD degrees in electrical engineering from the University of Wisconsin-Madison in 2005 and 2009, respectively. He was a postdoctoral research associate in the Program in Applied and Computational Mathematics at Princeton University from 2009 to 2010, and a research scientist in the Department of Electrical and Computer Engineering at Duke University from 2010 to 2011. He has been with Rutgers University since 2011, where he is currently an associate professor in the Department of Electrical and Computer Engineering and an associate member of the graduate faculty of the Department of Statistics. His research interests include statistical signal processing, high-dimensional statistics, machine learning, harmonic analysis, inverse problems, and networked systems.

Dr. Bajwa has received a number of awards in his career including the Army Research Office Young Investigator Award (2014), the National Science Foundation CAREER Award (2015), Rutgers University’s Presidential Merit Award (2016), Rutgers University’s Presidential Fellowship for Teaching Excellence (2017), and Rutgers Engineering Governing Council ECE Professor of the Year Award (2016, 2017, 2019). He is a co-investigator on a work that received the Cancer Institute of New Jersey’s Gallo Award for Scientific Excellence in 2017, a co-author on papers that received Best Student Paper Awards at IEEE IVMSP 2016 and IEEE CAMSAP 2017 workshops, and a Member of the Class of 2015 National Academy of Engineering Frontiers of Engineering Education Symposium.
\end{IEEEbiography}
 \vskip 0pt plus -1fil
\begin{IEEEbiographynophoto}
. He has also been involved in a number of professional activities. He served as the Lead Guest Editor for a special issue of IEEE Signal Processing Magazine on "Distributed, Streaming Machine Learning" (2020), a Publicity and Publications Co-Chair for IEEE DSW 2019 Workshop, the US Liaison Chair for IEEE SPAWC 2019 Workshop, Technical Co-Chair of IEEE SPAWC 2018 Workshop, a Technical Area Chair of 2018 Asilomar Conference on Signals, Systems, and Computers, the General Chair of 2017 DIMACS Workshop on Distributed Optimization, Information Processing, and Learning, an elected member of the Machine Learning for Signal Processing (MLSP) Technical Committee of the IEEE Signal Society (2016–2018), the Publicity and Publications Chair of IEEE CAMSAP 2015 Workshop, and an Associate Editor of IEEE Signal Processing Letters (2014–2017). He also co-chaired IEEE GlobalSIP 2013 Symposium on New Sensing and Statistical Inference Methods and CPSWeek 2013 Workshop on Signal Processing Advances in Sensor Networks, and co-guest edited a special issue of Elsevier Physical Communication Journal on "Compressive Sensing in Communications" (2012). He is currently serving as a Guest Editor for a special issue of Proceedings of the IEEE on "Optimization for Data-driven Learning and Control," a Senior Area Editor for IEEE Signal Processing Letters, an Associate Editor for IEEE Transactions on Signal and Information Processing over Networks, and an elected member of Sensor Array and Multichannel (SAM) and Signal Processing for Communications and Networking (SPCOM) Technical Committees of IEEE Signal Processing Society.
\end{IEEEbiographynophoto}
 \vskip 0pt plus -1fil
\begin{IEEEbiography}[{\includegraphics[width=1in,height=1.25in,clip,keepaspectratio]{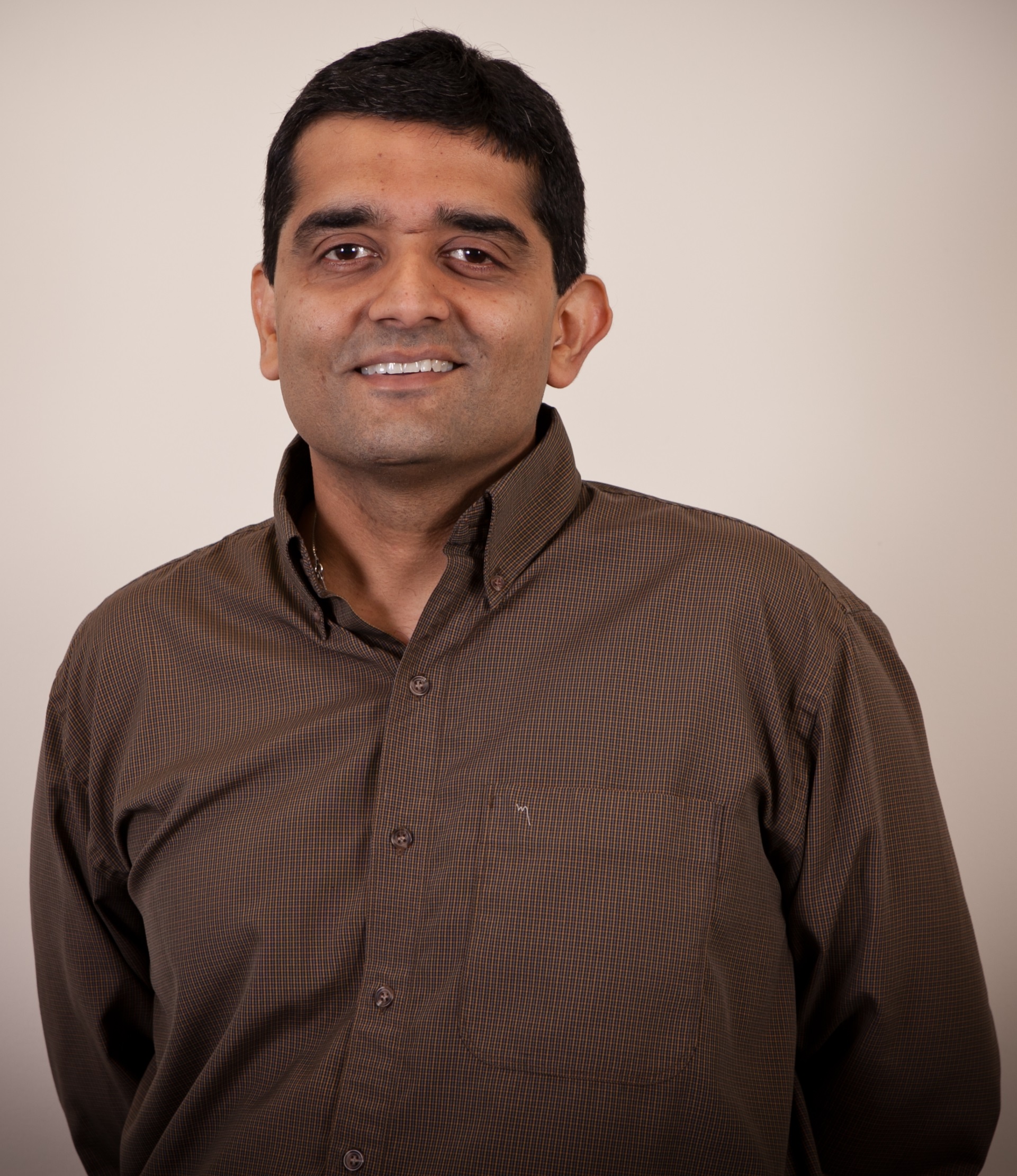}}]{Narayan B. Mandayam} (S’89-M’94-SM’99-F’09) received the B.Tech (Hons.) degree in 1989 from the Indian Institute of Technology, Kharagpur, and the M.S. and Ph.D. degrees in 1991 and 1994 from Rice University, all in electrical engineering.  Since 1994 he has been at Rutgers University where he is currently a Distinguished Professor and Chair of the Electrical and Computer Engineering department. He also serves as Associate Director at WINLAB. He was a visiting faculty fellow in the Department of Electrical Engineering, Princeton University, in 2002 and a visiting faculty at the Indian Institute of Science, Bangalore, India in 2003. Using constructs from game theory, communications and networking, his work has focused on system modeling, information processing and resource management for enabling cognitive wireless technologies to support various applications. His recent interests include privacy in IoT, resilient smart cities as well as modeling and analysis of trustworthy knowledge creation on the internet. He has also been working recently on the use of prospect theory in understanding the psychophysics of pricing for wireless data networks as well as the smart grid.

Dr. Mandayam is a co-recipient of the 2015 IEEE Communications Society Advances in Communications Award for his seminal work on power control and pricing, the 2014 IEEE Donald G. Fink Award for his IEEE Proceedings paper titled “Frontiers of Wireless and Mobile Communications” and the 2009 Fred W. Ellersick Prize from the IEEE Communications Society for his work on dynamic spectrum access models and spectrum policy. He is also a recipient of the Peter D. Cherasia Faculty Scholar Award from Rutgers University (2010), the National Science Foundation CAREER Award (1998), the Institute Silver Medal from the Indian Institute of Technology (1989) and its Distinguished Alumnus Award (2018). He is a coauthor of the books: Principles of Cognitive Radio (Cambridge University Press, 2012) and Wireless Networks: Multiuser Detection in Cross-Layer Design (Springer, 2004). He has served as an Editor for the journals IEEE Communication Letters and IEEE Transactions on Wireless Communications. He has also served as a guest editor of the IEEE JSAC Special Issues on Adaptive, Spectrum Agile and Cognitive Radio Networks (2007) and Game Theory in Communication Systems (2008). He is a Fellow and Distinguished Lecturer of the IEEE.
\end{IEEEbiography}

\end{document}